\documentclass[10pt,twocolumn,letterpaper]{article}

\usepackage{cvpr}              % To produce the CAMERA-READY version
\usepackage{graphicx}
\usepackage{amsmath}
\usepackage{amssymb}
\usepackage{booktabs}
\usepackage{array}
\usepackage{tabularx}
\usepackage{pifont}% http://ctan.org/pkg/pifont
\usepackage{color, colortbl}

\usepackage{booktabs, multirow, makecell} % for borders and merged ranges
\usepackage{soul}% for underlines
\usepackage[table]{xcolor} % for cell colors

\usepackage{amssymb}% http://ctan.org/pkg/amssymb
\usepackage{pifont}% http://ctan.org/pkg/pifont
\newcommand{\cmark}{\checkmark}%
\newcommand{\xmark}{}%

\newcommand{\ospatrack}{$\mathcal{O}_{pose}^2$}
\newcommand{\ospa}{$\mathcal{O}_{pose}$ }

\definecolor{Gray}{gray}{0.9}

\usepackage[pagebackref,breaklinks,colorlinks]{hyperref}

\usepackage[capitalize]{cleveref}
\crefname{section}{Sec.}{Secs.}
\Crefname{section}{Section}{Sections}
\crefname{table}{Tab.}{Tabs.}
\Crefname{table}{Table}{Tables}

\newcommand{\tho}[1][black]{\textcolor{#1}}
\newcommand{\eddie}[1][black]{\textcolor{#1}}

\def\cvprPaperID{7077} % *** Enter the CVPR Paper ID here
\def\confName{CVPR}
\def\confYear{2022}

\title{JRDB-Pose: A Large-scale Dataset for Multi-Person Pose Estimation and Tracking}

\author{Edward Vendrow$^{1}$\thanks{equal contribution}\ , Duy Tho Le$^{2}$\footnotemark[1]\ , Jianfei Cai$^{2}$, Hamid Rezatofighi$^{2}$\\
$^{1}$Stanford University, $^{2}$Monash University \\ 
{\tt\small evendrow@stanford.edu, \{tho.le, hamid.rezatofighi\}@monash.edu}
}
\begin{document}
% \twocolumn[{%
% \renewcommand\twocolumn[1][]{#1}%
% \maketitle
% \begin{center}
%     \centering
%     \captionsetup{type=figure}
%     \includegraphics[width=.9\textwidth]{latex/figures/pose1.png}
%     \captionof{figure}{Test caption}
% \end{center}%
% }]
\makeatother
\maketitle

%%%%%%%%% ABSTRACT
\begin{abstract}
Autonomous robotic systems operating in human environments must understand their surroundings to make accurate and safe decisions. In crowded human scenes with close-up human-robot interaction and robot navigation, a deep understanding of surrounding people requires reasoning about human motion and body dynamics over time with human body pose estimation and tracking. However, existing datasets captured from robot platforms either do not provide pose annotations or do not reflect the scene distribution of social robots. In this paper, we introduce JRDB-Pose, a large-scale dataset and benchmark for multi-person pose estimation and tracking. JRDB-Pose extends the existing JRDB which includes videos captured from a social navigation robot in a university campus environment, containing challenging scenes with crowded indoor and outdoor locations and a diverse range of scales and occlusion types. JRDB-Pose provides human pose annotations with per-keypoint occlusion labels and track IDs consistent across the scene and with existing annotations in JRDB. We conduct a thorough experimental study of state-of-the-art multi-person pose estimation and tracking methods on JRDB-Pose, showing that our dataset imposes new challenges for the existing methods. JRDB-Pose is available at \url{https://jrdb.erc.monash.edu/}.
\end{abstract}

%%%%%%%%% BODY TEXT
\section{Introduction}
\label{sec:intro}

\begin{table*}[ht]
% \small
\footnotesize
    \centering
%    \footnotesize
    \begin{tabular}{c|cccccccp{11mm}p{13mm}p{9mm}p{9mm}}
        \hline
Dataset                                             & \# Poses  & \# Boxes  & Tracks  & Crowd   & ppF & Occlusion & Action & Indoor + \newline Outdoor & Robot \newline Navigation & Multi- \newline Modal & Multi- \newline Task  \\ [0.5ex] \hline
% 3DPW \cite{von2018recovering}                       & \todo{?}       & -         & \xmark    & \xmark    & \xmark & 1-2 & \todo{?} & \xmark & \cmark  \\
MPII \cite{andriluka14mpii}                         & 40K       &          & \xmark    & \xmark    & 1-17 & \xmark & \xmark & \cmark & \xmark & \xmark & \xmark  \\
Penn Action \cite{zhang2013pennaction}              & 160k      & 160k      & \xmark    & \xmark    & 1 & \xmark & \cmark  & \cmark & \xmark & \xmark & \xmark  \\
COCO \cite{lin2014microsoftcoco}                    & 250k      & 500k      & \xmark    & \xmark    & \underline{1-20} & \cmark & \xmark & \cmark & \xmark & \xmark & \cmark \\
\hline
KITTI \cite{geiger2012kitti}                        &           & 80k       & \cmark    & \cmark    &       & \cmark & \xmark & \xmark  & \xmark & \cmark & \cmark \\
H3D \cite{patil2019h3d}                             &           & 460k      & \cmark    & \cmark    &       & \xmark & \xmark & \xmark  & \xmark & \cmark & \xmark \\
MOT20 \cite{dendorfer2020mot20}                     &           & 1.65M     & \cmark    & \cmark    &       & \cmark & \xmark & \cmark  & \xmark & \xmark & \xmark \\
THÖR \cite{rudenko2020thor}                         &           & 2.5M      & \cmark    & \xmark    &       & \xmark & \xmark & \xmark  & \cmark & \cmark & \xmark \\
\hline
PoseTrack21 \cite{doering2022posetrack21}           &  \underline{177k}     & 429k      & \cmark    & \cmark    & 1-13   & \cmark & \xmark & \cmark  & \xmark & \xmark & \xmark \\
Waymo \cite{sun2020scalabilitywaymo}                & 173K      & \textbf{9.9M}      & \cmark    & \cmark    & \textit{unk}    & \cmark & \xmark & \xmark  & \xmark & \cmark & \cmark \\
\hline
\textbf{JRDB-Pose}                                  & \textbf{636k}      & \underline{2.8M}      & \cmark    & \cmark    & \textbf{1-36}  & \cmark & \cmark & \cmark & \cmark & \cmark & \cmark \\ \hline
\hline
\rowcolor{Gray}
JTA$^\dagger$ \cite{fabbri2021motsynth}                   &  10M      &           & \xmark    & \cmark    & 0-60   & \cmark & \xmark & \cmark & \xmark & \xmark & \xmark \\
\rowcolor{Gray}
MotSynth$^\dagger$ \cite{fabbri2021motsynth}              &  40M      &           & \cmark    & \cmark    & 0-125  & \cmark & \xmark & \cmark & \xmark & \xmark & \xmark \\\hline\hline

    \end{tabular}\vspace{-.5em}
    \caption{Comparison of existing public datasets related to 2D pose estimation and tracking. For each dataset we report the numbers of poses, boxes, as well as the availability of tracking information, crowd data, people per frame (ppF), occlusion labels, action labels, scene type, and if the data comes from robot navigation in human environments. We mark if a dataset has data modalities besides RGB frames, and if it contains annotations for multi-task types. Note that JRDB-Pose is a multi-modal dataset captured from a social navigation robot, addressing different research challenges than many existing works.  \ \ \ \     \textit{$^\dagger$Synthetic dataset.}\ \ \ \     \textit{unk: Unknown.}
    % \hrtc{make all tables small or footnote size}
    }
    \label{tab:comparison}
    \vspace{-1.5em}
\end{table*}

Visual scene understanding of human environments is a difficult and crucial task for autonomous driving, human-robot interaction, safe robotic navigation, and human action recognition. Although rough predictions of human location are sufficient for some applications, a deep understanding of crowded human scenes and close-up human-robot interaction requires reasoning about human motion and body dynamics with human body pose estimation and tracking. Developing an AI model to predict human body pose is made more difficult by the varied and highly imbalanced range of human motion found in daily living environments, including a variety of scales, occlusions, and overlapping humans, representing a long-tailed distribution of human poses which is difficult for existing methods.

\begin{figure}
    \centering
    
    \begin{minipage}[h]{1.0\linewidth}
    \centering
    \includegraphics[width=\linewidth]{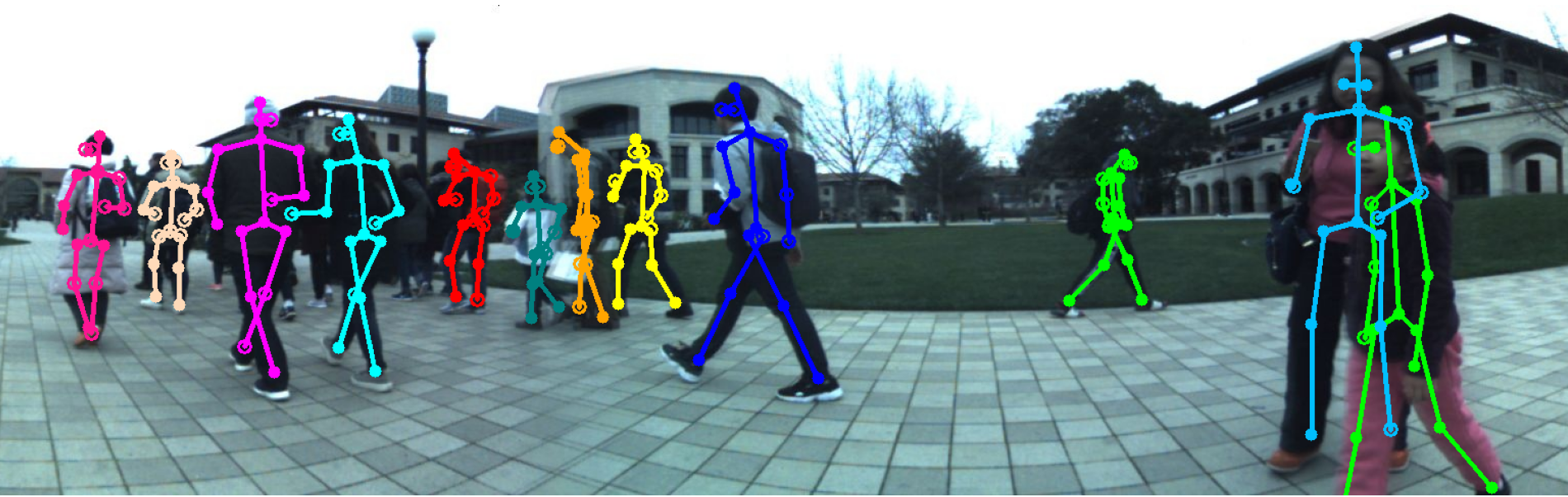}
    \end{minipage}
    
    \vspace{0.00mm} 

    %second figure 
    \begin{minipage}[h]{1.0\linewidth}
    \centering
    \includegraphics[width=\linewidth]{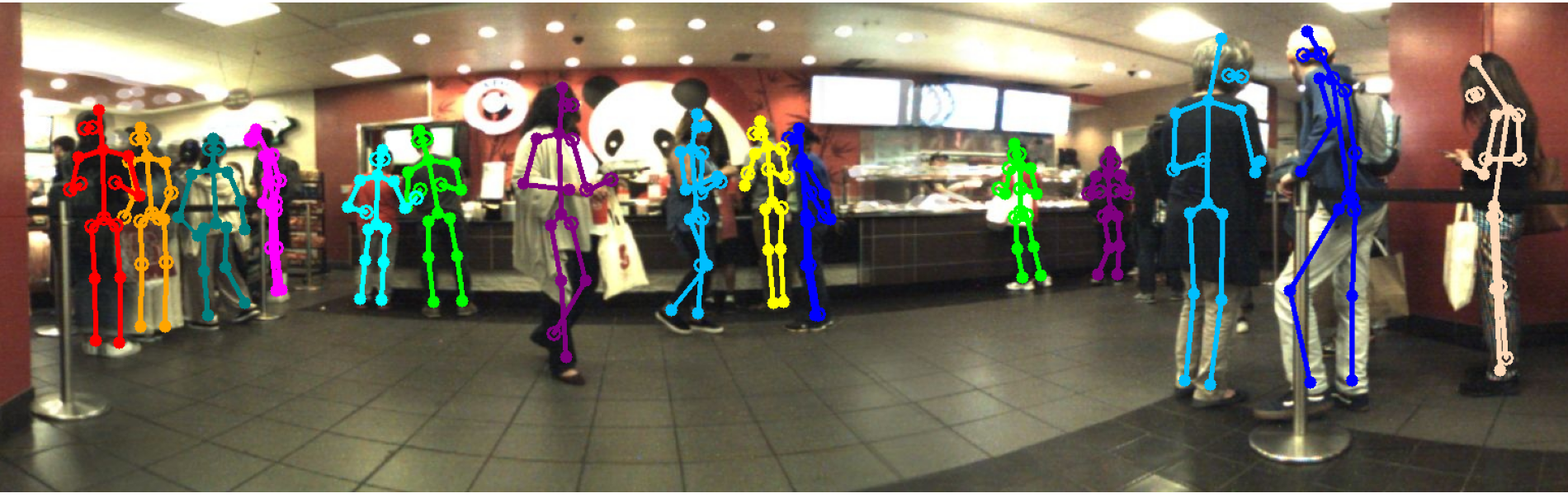}
    \end{minipage}  
\vspace{-.1em}
    \caption{JRDB-Pose provides high frequency annotations of tracks and body joints in long scenes of crowded indoor and outdoor locations featuring dynamic motion and occlusion. }
    \label{fig:pose_example}
    \vspace{-1.5em}
\end{figure}

Human pose estimation and tracking is an active research area with many new large-scale datasets \cite{doering2022posetrack21, sun2020scalabilitywaymo, lin2014microsoftcoco, fabbri2021motsynth} contributing to significant recent progress; however, these datasets do not primarily target robotic perception tasks in social navigation environments, and thus rarely reflect specific challenges found in human-robot interaction and robot navigation in crowded human environments, \eg shopping malls, university campus, \etc.
% \hrtc{in order to avoid a disagreement for the reader, support it with this statement that you will show these unique challenges in this paper which challenges existing SOTA models}

JRDB \cite{martin2021jrdb} previously introduced a large-scale dataset and a benchmark for research in perception tasks related to robotics in human environments. The dataset was captured using a social manipulator robot with a multi-modal sensor suite including a stereo RGB 360° cylindrical video stream, 3D point clouds from two LiDAR sensors, audio and GPS positions. JRDB \cite{martin2021jrdb} additionally introduced annotations for 2D bounding boxes and 3D oriented cuboids. Recently, JRDB-Act \cite{ehsanpour2022jrdb} further introduced new annotations on the JRDB videos for individual actions, human social group formation, and social activity of each social group. 
%We extend this valuable dataset by providing human body pose annotation.   

JRDB was collected from a robotic navigation platform in crowded human environments, diversely capturing both indoor and outdoor scenes. Additionally since the robot's camera is located at person-level, and moves around, the data is not just collected from a far-off view but captures close-up scenes. 

For robotic systems to safely navigate dynamic human environments and perform collision risk prediction, they must be able to accurately track and forecast motion of people in their surroundings. Human motion is often fast and requires high frame rate data for accurate prediction and tracking, making high-frequency annotated human pose data crucial for the development and evaluation of robotic perception systems in human environments. \eddie{Complex social interactions add difficulty and similarly benefit from high-frequency data.} In crowded scenes with high levels of occlusions or overlap with other humans, tracking may be also difficult.

We introduce JRDB-Pose, a large-scale dataset captured from a mobile robot platform containing human pose and head box annotations. JRDB-Pose provides 600k pose annotations and 600k head box annotations, each with an associated tracking ID. JRDB-Pose includes a wide distribution of pose scales and occlusion levels, each with per-keypoint occlusion labels and consistent tracking IDs across periods of occlusion. The combination of JRDB-Pose with JRDB and JRDB-Act forms a valuable multi-modal dataset providing a comprehensive suite of annotations suited for robotic interaction and navigation tasks. 
\\
Our contributions are:\vspace{-.5em}
\begin{itemize}
	\setlength{\itemsep}{1pt}
	\setlength{\parskip}{0pt}
	\setlength{\parsep}{0pt}
    \item We introduce JRDB-Pose, a large-scale pose estimation and tracking dataset providing pose annotations and head boxes with tracking IDs and per-keypoint occlusion labels.
    % \item JRDB-Pose annotates a 17-keypoint skeleton with per-keypoint occlusion severity labels.
    % \item JRDB-Pose contains challenge scenes with crowded indoor and outdoor locations and a diverse range of scales and occlusion types.
    % \item We introduce two challenging benchmarks for pose estimation and pose tracking with a public evaluation server on a held-out test set \hrtc{Please don't emphesize on ``a public evaluation server on a held-out test set'' as it may violate anonymous policy.  just highlight the we introduce a new benchmark with new metrics in addition to popular one, which is the next contribtuion.}.
    \item In addition to adopting the popular metrics, we introduce new metrics, OSPA-Pose and OSPA$^{(2)}$-Pose for pose estimation and tracking, respectively.
    \item We conduct a comprehensive evaluation of state-of-the-art methods on JRDB-Pose and discuss the strengths and weaknesses of existing methods.
\end{itemize}

%%%%%%%%% RELATED DATASETS
\section{Related Datasets}

% \hrtc{Add discussion of MOT tracking papers with many citations. 1.2-1.5 column to datasets and 0.5-0.8 column to SOTA methods in pose estimation and tracking} 
We summarize the commonly used public datasets for pose estimation \cite{von20183dpw, zhang2013pennaction, lin2014microsoftcoco, andriluka14mpii, lin2014microsoftcoco}, pose tracking \cite{doering2022posetrack21, sun2020scalabilitywaymo} and multi-object tracking \cite{geiger2012kitti, dendorfer2020mot20, patil2019h3d, rudenko2020thor} in \cref{tab:comparison}.

\begin{figure*}[ht]
    \centering
    \includegraphics[height=80pt]{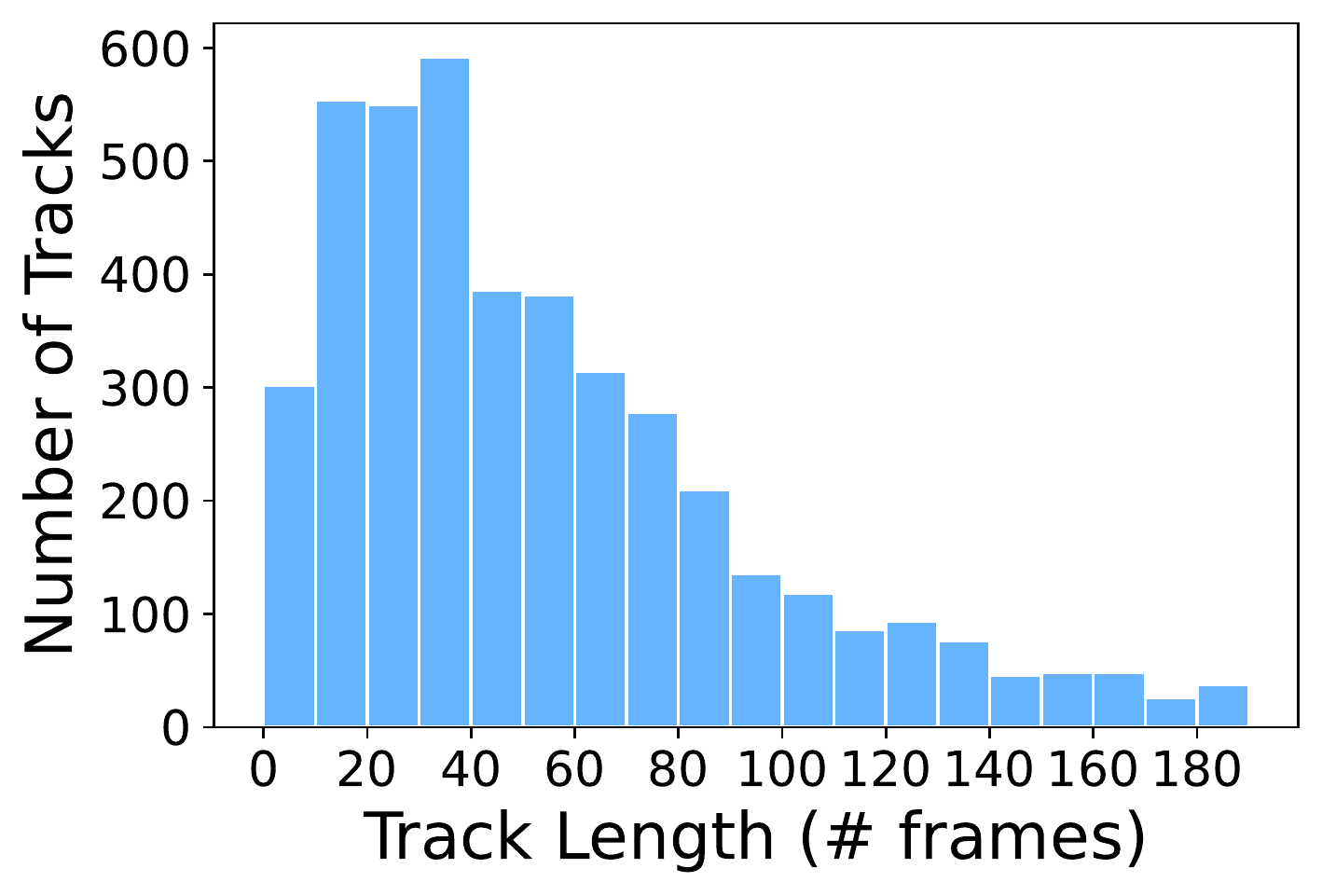}
    \includegraphics[height=80pt]{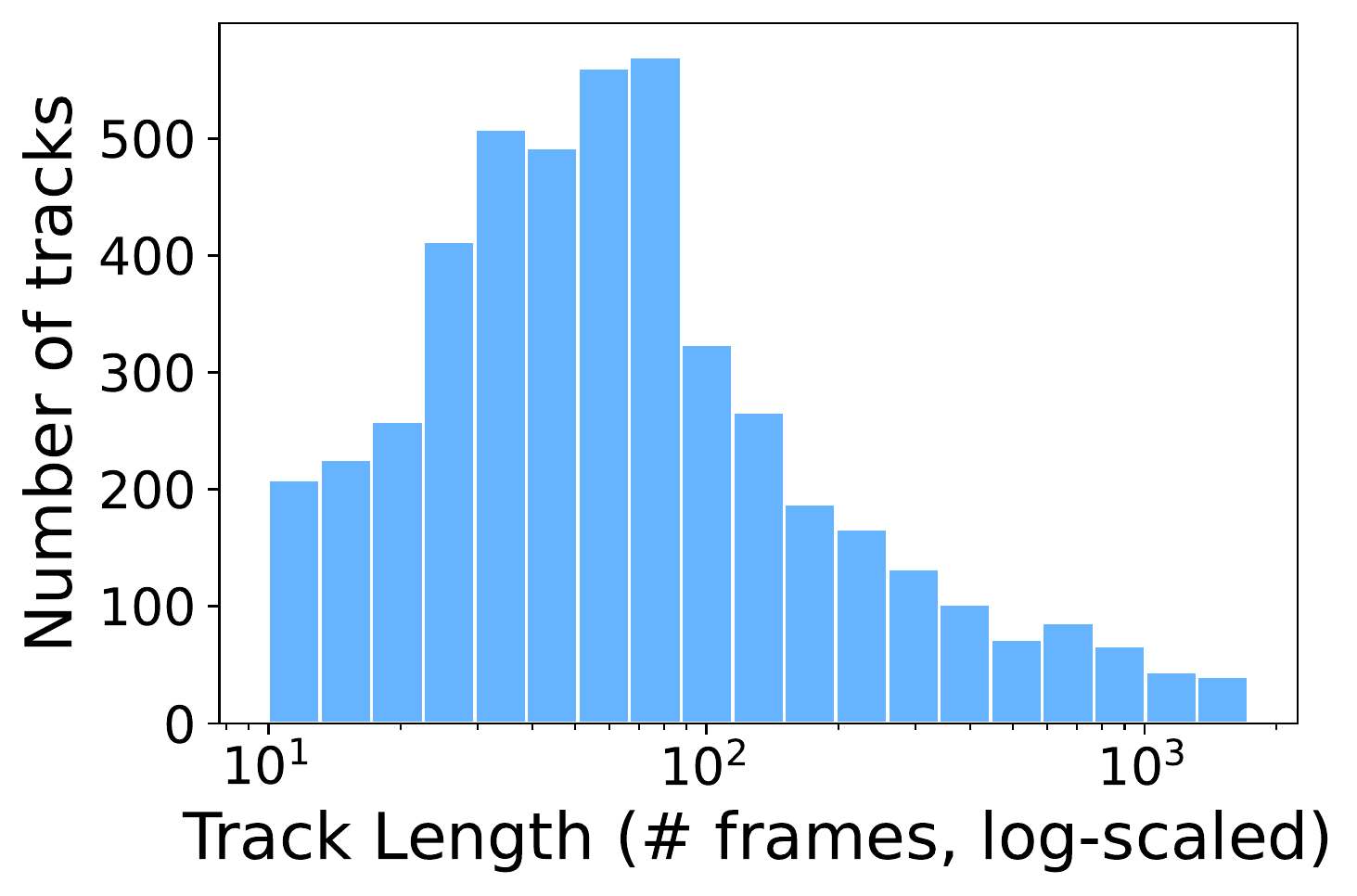}
    \includegraphics[height=80pt]{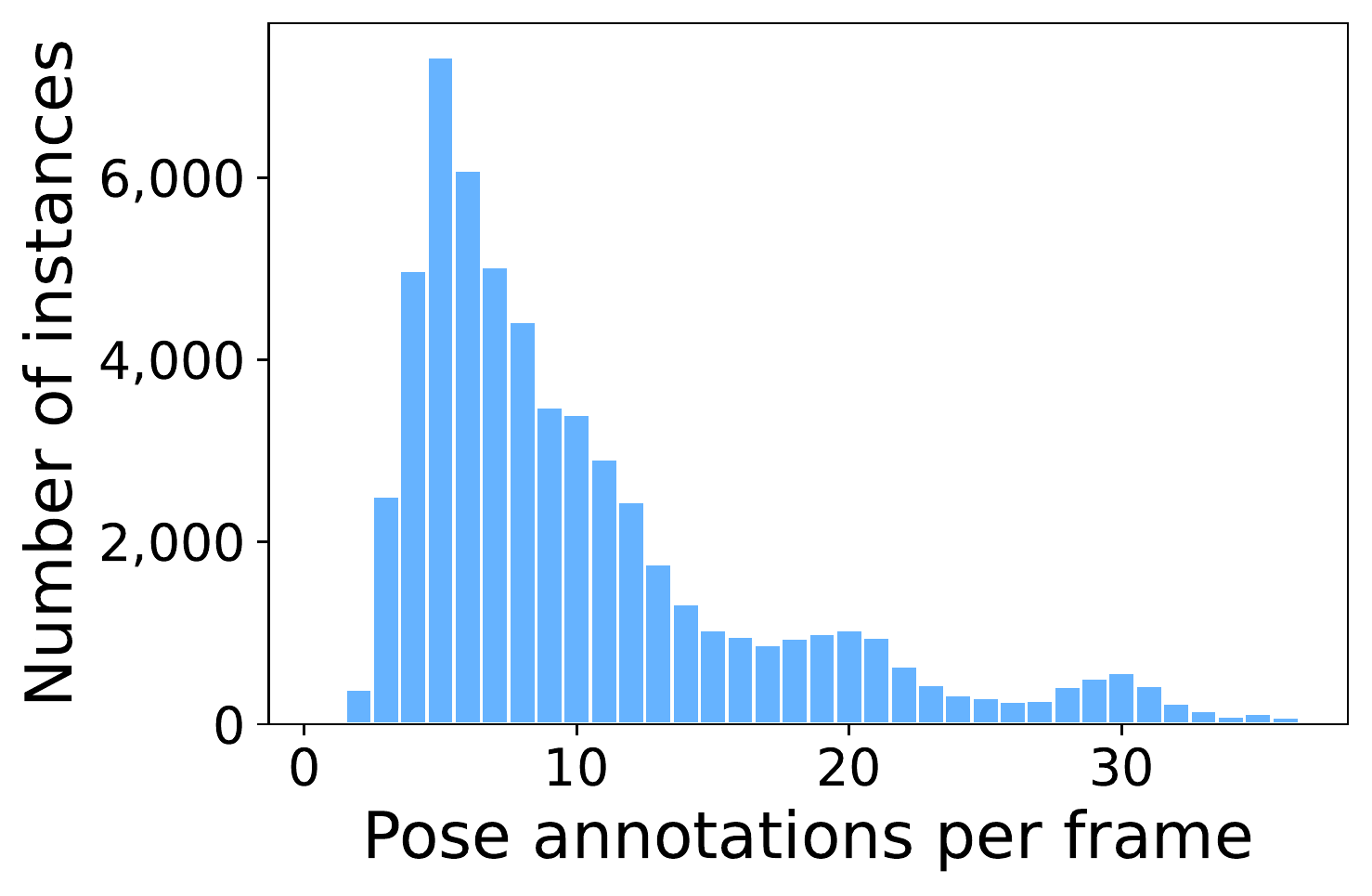}
    \includegraphics[height=80pt]{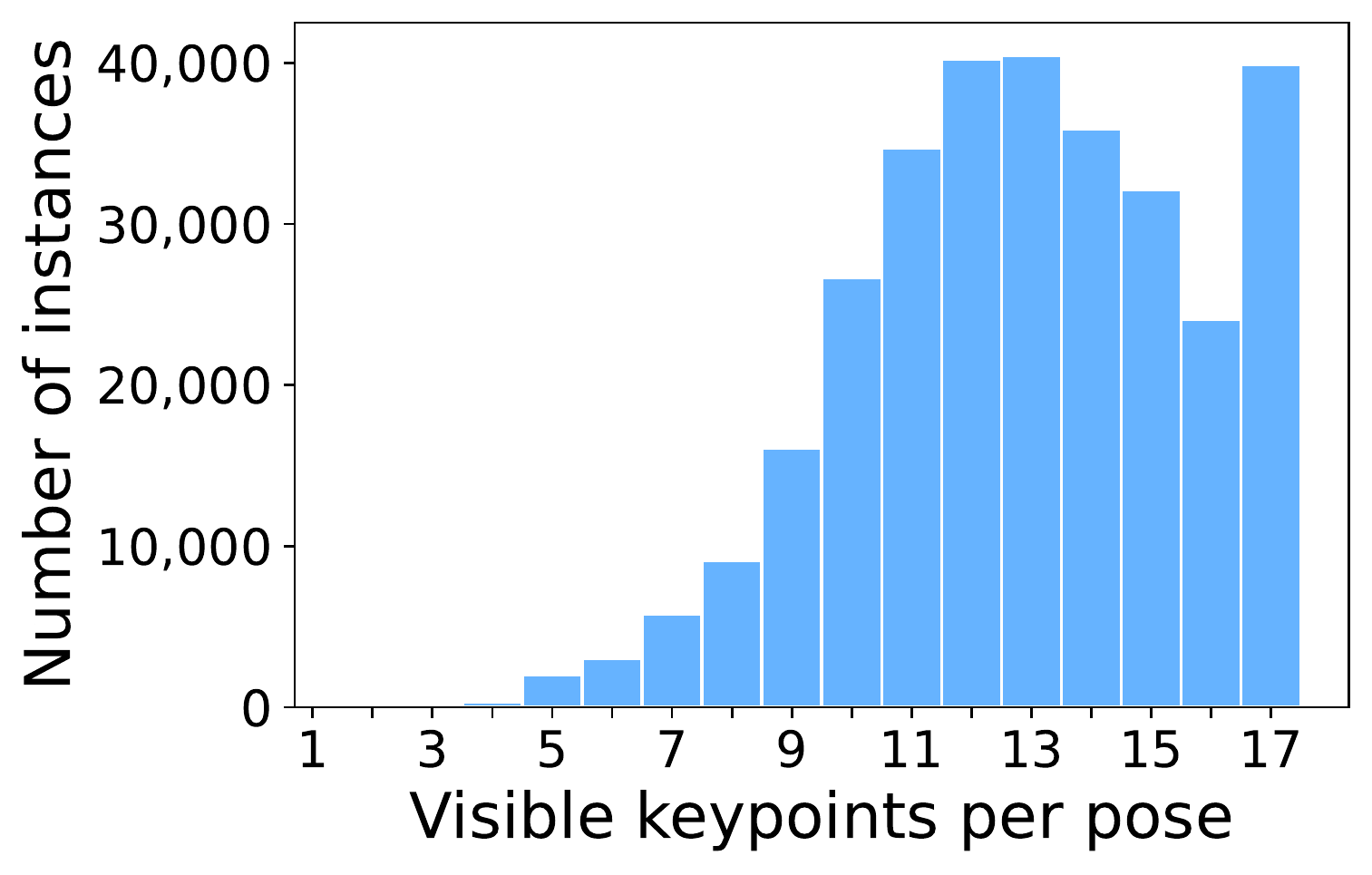}\vspace{-.5em}
    \caption{
    % \hrtc{TODO: Increase both axis font size}
    Various statistic for JRDB-Pose, which provides visibility labels and track ids consistent across long periods of occlusion. From left to right: \textit{1)} The distribution of track lengths, with the long tail truncated. \textit{2)} A log-scaled distribution showing all track lengths. JRDB-Pose track lengths are varied and as high as 1700 frames long. \textit{3)} Number of pose annotations in each panoramic frame. \textit{4)} Number of keypoints in each pose annotated as visible. }
    \label{fig:track_length}
    \vspace{-1em}
\end{figure*}
\begin{figure}
    \centering
    \begin{subfigure}[h]{.45\linewidth}
    \centering
    \includegraphics[width=\textwidth]{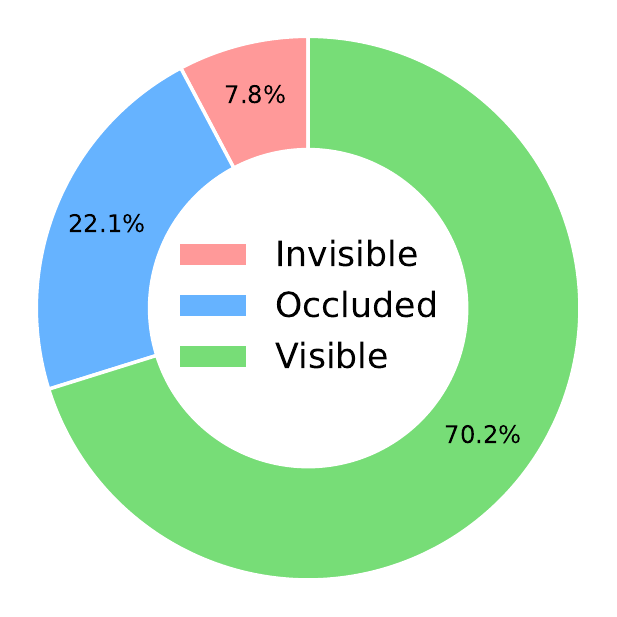}
    \caption{Train set}
    \end{subfigure}
    \hfill
    \begin{subfigure}[h]{.45\linewidth}
    \centering
    \includegraphics[width=\textwidth]{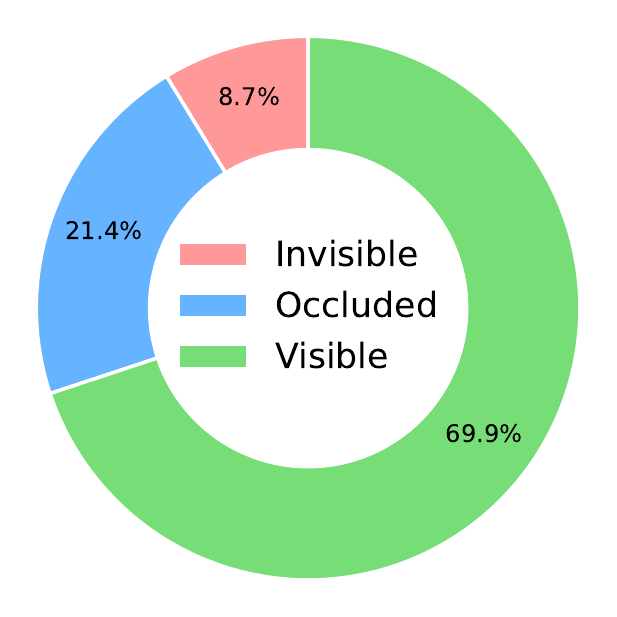}
    \caption{Test set}
    \end{subfigure}\vspace{-.5em}
    \caption{Distribution of keypoint visibility annotations in the JRDB-Pose train/validation and test splits. While most joints are visible, JRDB-Pose contains a large number of occluded and invisible joints.} \vspace{-1em}
    \label{fig:vis_train_test}
\end{figure}

\noindent
\textbf{Single and Multi-person Pose Estimation Datasets:}
The task of 2D pose estimation involves predicting the pixel locations of human skeleton keypoints on an image. The MPII Human Pose Dataset \cite{andriluka14mpii} is a popular multi-person pose estimation dataset and benchmark including videos from everyday human activities, and the larger Penn Action dataset \cite{zhang2013pennaction} provides both pose and action annotations in sports settings with a single pose per frame. MS COCO Keypoints Challenge \cite{lin2014microsoftcoco} proposed a large-scale dataset including a diverse set of scenes and occlusion labels. All of these datasets label individual frames, limiting them to single-frame pose estimation. 
% \hrtc{I highlight temporal fps for our dataset annotation in Table 1 and all over the texts. you should be able to show it is very challenging to provide human body pose annotation in 2D with such high temporal resolution}. 
Human3.6M \cite{h36m_pami} annotates videos and single-person 3D poses from a controlled indoor scene, while 3DPW \cite{von20183dpw} provides estimated 3D human meshes in-the-wild with up to 2 people per frames. In comparison, JRDB-Pose is captured from a robotic platform in-the-wild with crowded and manually-annotated indoor and outdoor scenes, addressesing a different set of challenges. JRDB-Pose also includes diverse data modalities including cylindrical video, LiDAR point clouds, and rosbags.

\noindent
\textbf{Multi-Person and Multi-Object Tracking Datasets:}
% Miracle \cite{yu2018multi} uses a simple IOU tracker on top of a robust single-image keypoints detector in conjunction with a cascaded pyramid network. More recently, 
Multi-Person Pose Tracking (MPPT) and Multi-Object Tracking (MOT) are crucial tasks in robotic perception and navigation, where the challenge is to track, across a video, the body keypoints of individuals or the locations of objects,  respectively. MOT has attracted significant attention from the community with large-scale challenges~\cite{motchallenge15, motchallenge16, dendorfer2020mot20}, and MPPT is increasingly becoming recognized for its significance in human activity understanding and human-object interaction. Despite high performance on easy scenes, current MPPT methods struggle in crowded environments with occlusions and scale fluctuations. 

Posetrack 2018 \cite{andriluka2018posetrack} introduced a benchmark for pose estimation and tracking using annotated video sequences featuring a variety of in-the-wild scenarios such as sports and dancing. Since PoseTrack was not densely labeled in crowded scenes, it required large ``ignore regions'' contains regions of images and videos lacking annotations. More recently, Posetrack21 \cite{doering2022posetrack21} used the videos from PoseTrack 2018 to provide dense annotations of crowded scenes. Both datasets target diverse pose tracking scenarios, containing videos such as surveillance-like footage that are different from typical robotic navigation and human-robot interaction scenarios. Additionally, each video sequence is limited to five seconds, while sequences in JRDB-Pose are up to 2 minutes long. JRDB-Pose is a challenging dataset and benchmark for real-world tracking tasks, containing videos captured in crowded environments with humans fading into and emerging from the crowd.

\begin{figure*}
    \centering
    \includegraphics[width=\linewidth]{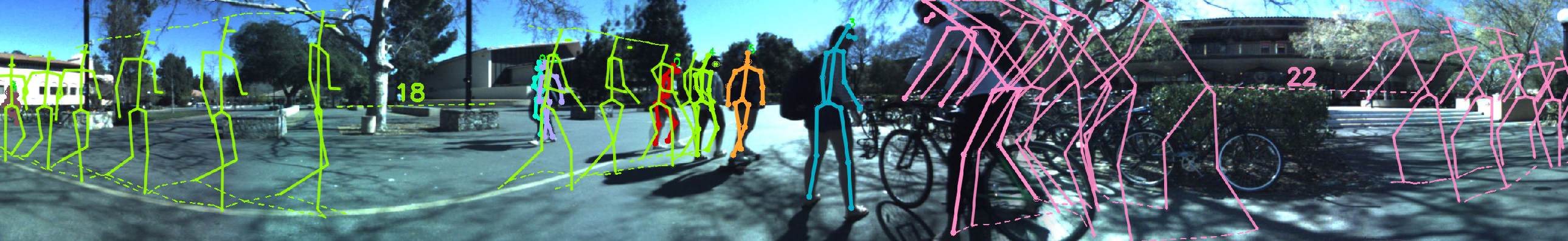}\vspace{-.5em}
    \caption{A portion of the panoramic frame with annotated pose instances. We show some of our tracking annotations by visualizing pose histories for two people. The gaps in the pose history correspond to periods of occlusion denoted by the numbers marking the length of the occlusion in frames. JRDB-Pose provides track IDs which are consistent even across long periods of occlusion.}
    \label{fig:pose_track_vis}
    \vspace{-1.0em}
\end{figure*}

\noindent
\textbf{Autonomous Driving Datasets:} Recently, autonomous vehicle datasets have been released featuring large-scale and detailed annotations of the surrounding environment, posing difficult MOT and MPPT challenges. H3D \cite{patil2019h3d} annotates human detections and tracks. More recently, the Waymo Open Dataset \cite{sun2020scalabilitywaymo} filled the gap by providing 172.6K human pose annotations from their autonomous vehicle navigating outdoor road and highway environments. While the data is valuable for driving applications, the dataset contains exclusively outdoor road and highway environments making it unsuitable for non-driving robotic tasks, \eg navigating a crowded shopping mall.

\noindent
\textbf{Robotic Navigation:}
Robotics operating in a social environment must learn to navigate in a socially-compliant manner by detecting and reacting to other agents in the environment. Existing datasets for robotic navigation \cite{martin2021jrdb, rudenko2020thor, karnan2022scand} provide relatively few annotations and are often created in limited environment. The THÖR dataset provides human trajectories on a ground plane, and is filmed in a controlled indoor environment. SCAND \cite{karnan2022scand} captures a variety of crowded indoor and outdoor environments without annotations for human detection or tracking. In human environments with complex interactions, it is crucial to understand human motion more deeply than a bounding box. To this end, JRDB-Pose is one of the first datasets providing large-scale pose annotations in a robotic navigation environment.

\noindent
\textbf{Synthetic Datasets:} Several recent works have proposed large synthetic datasets for human pose estimation and tracking by generating data using a video game engine. The Joint Track Auto (JTA) \cite{fabbri2018jta} dataset generates 10M annotated human body poses, which MotSynth \cite{fabbri2021motsynth} extends to 40M human poses with associated track IDs. While synthetic datasets have been shown to be helpful to the existing frameworks' performance boost when combined with the real-world datasets~\cite{fabbri2021motsynth}, they may not reflect underlying biases and distributions in real-world data, preventing them from being the only training data resources for evaluating approaches in real-world applications.
\\
\eddie{\textbf{State-of-the-art Frameworks:} Multi-person pose estimation involves predicting body keypoint locations for all people in an image and identifying which keypoints belong to which individuals. There are two common types of approaches to this problem: top-down methods first detect all people and then execute pose estimation for each individual~\cite{wang2020deep, khirodkar2021multi, ning2018top, yu2018multi}, while bottom-up methods first identify keypoints directly and then group them into individual people~\cite{braso2021center}, including disentangled representations~\cite{geng2021bottom}, associative embedding~\cite{newell2017associative}, HGG~\cite{jin2020differentiable}, YOLO-Pose~\cite{maji2022yolo}, and CID~\cite{wang2022contextualcid}}.

\tho{Multi-person pose tracking approaches \cite{ning2018top, yu2018multi, braso2021center, snower202015} often identify keypoints or poses using a pose estimation method, and then predict tracks using the estimated keypoints. Top-down techniques like  openSVAI \cite{ning2018top} and PGG \cite{jin2019multi} decompose the task into three discrete stages: human detection, pose estimation, and then pose tracking. Recently, UniTrack \cite{unitrack} proposed a unified framework for multiple object and keypoint tracking utilizing shared general-purpose appearance models.}

\tho{The multi-object tracking task can be categorized into tracking by detection (TBD)~\cite{bytetrack,deepsort, ocsort, strongsort,motioninfo3} and joint detection and tracking (JDT)~\cite{tracktor, centertrack, sun2020transtrack, trackformer, transcenter, transmot,lookingbeyond}. SORT-based\cite{sort} approaches \cite{deepsort, ocsort, strongsort} have gained popularity with methods including DeepSORT \cite{deepsort} 
% which embeds cosine distance together with extra appearance vectors while
and StrongSORT\cite{strongsort}
% uses AFLink and GSI modules to associate short trackets and compensate false negative detections.
In addition, OC-Sort \cite{ocsort} improved these methods by proposing an enhanced filter and recovery strategy suitable for tracking non-linearly moving and frequently occluded objects. Recently, ByteTrack \cite{bytetrack} achieved efficient and precise tracking results using a general association technique that utilizes nearly every detection box. Popular JDT methods include Tracktor++\cite{tracktor} which uses integrated detection-tracking modules and CenterTrack\cite{centertrack} which employs two-frames tracking algorithm for real-time and accurate tracking. Overall, since significant research in multi-person pose and multi-object tracking focuses on refining track consistency, occlusion recovery, and missing and false detection handling, JRDB-Pose is a useful and challenging benchmark for existing and future works. }
% More recently, Transformer-based methods \cite{sun2020transtrack, trackformer, transcenter, transmot, lookingbeyond} leverage the Query-Key mechanism using Spatial and Temporal Transformers for combined detection and tracking, showing tremendous success in multi-object tracking field.}

% The multi-object tracking algorithms are divided into
% TBD, JDT, and Transformer-based tracking methods, based on the outcomes of data catego-
% rization and the various network structures of the tracking algorithms

%%%%%%%%% JRDBPOSE DATASET
\section{The JRDB-Pose Dataset}

We create JRDB-Pose using all videos from the JRDB dataset \cite{martin2021jrdb}, including 64 minutes of sensory data consisting of 54 sequences captured from indoor and outdoor locations in a university campus environment. JRDB-Pose includes 57,687 annotated panoramic (3760 x 480) frames, containing total of 636k annotated pose instances with 11 million labeled keypoints with occlusion labels for each keypoint. \eddie{We annotate 17 keypoints for each body pose including head, eyes, neck, shoulders, center-shoulder, elbows, wrists, hips, center-hips, knees and ankles}. Each pose includes a tracking ID which is consistent for each person across the entire scene, including across long periods of full occlusion (see Fig.~\ref{fig:pose_track_vis})\eddie{, and is also consistent with person IDs for existing 2D and 3D bounding box annotations from JRDB}. Fig.~\ref{fig:track_length} shows the distribution of track lengths present in JRDB-Pose. In total, JRDB-Pose provides 5022 pose tracks 
% \hrtc{How we have more tracks compared to bounding boxes?}
with an average length of 124 frames and some tracks as long as 1700 frames. 

\begin{figure*}
    \centering
    \includegraphics[width=\textwidth]{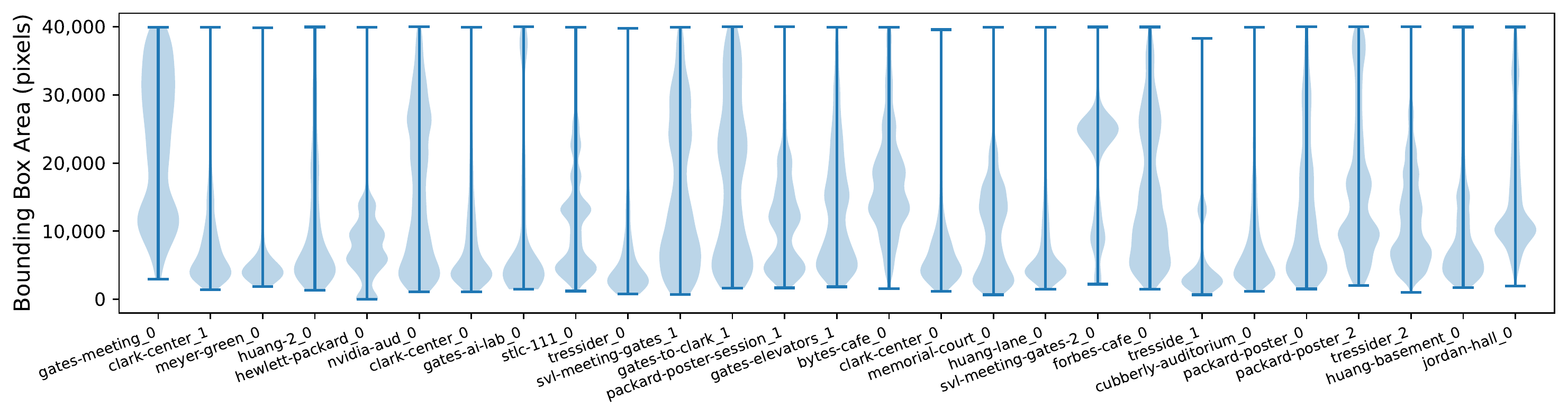}\vspace{-.5em}
    \caption{Distribution of bounding box scales in the train and validation scenes. JRDB-Pose contains a wide distribution of pose scale that is different for each scene.}
    \label{fig:scale_by_scene}
    \vspace{-1em}
\end{figure*}
\begin{table}
\small
    \centering
    \begin{tabular}{c|c|c}
    \hline
        Annotations & Quantity & Track IDs \\ [0.5ex] \hline
        2D \tho{Body} Bounding Box \cite{martin2021jrdb} & 2,400,000 & \cmark  \\
        3D \tho{Body} Bounding Box \cite{martin2021jrdb} & 1,800,000 & \cmark \\
        Atomic Action \cite{ehsanpour2022jrdb} & 2,800,000 & N/A \\
        Social Group \cite{ehsanpour2022jrdb} & 600,000 & N/A \\
        \textbf{2D Head Bounding Box} & 600,000 & \cmark \\ 
        \textbf{2D Body Pose} & 600,000 & \cmark \\ \hline
    \end{tabular} \vspace{-.5em}
    \caption{A summary of annotations in by JRDB \cite{martin2021jrdb}, JRDB-Act \cite{ehsanpour2022jrdb}, and JRDB-Pose, which together provide a unique multi-modal dataset suitable for multi-person detection, pose estimation tracking, and activity recognition. Bolded annotations are introduced in this paper. } \vspace{-.5em}
    \label{tab:jrdb_overview}
\end{table}
\begin{figure}
    \centering
    \includegraphics[width=0.9\linewidth]{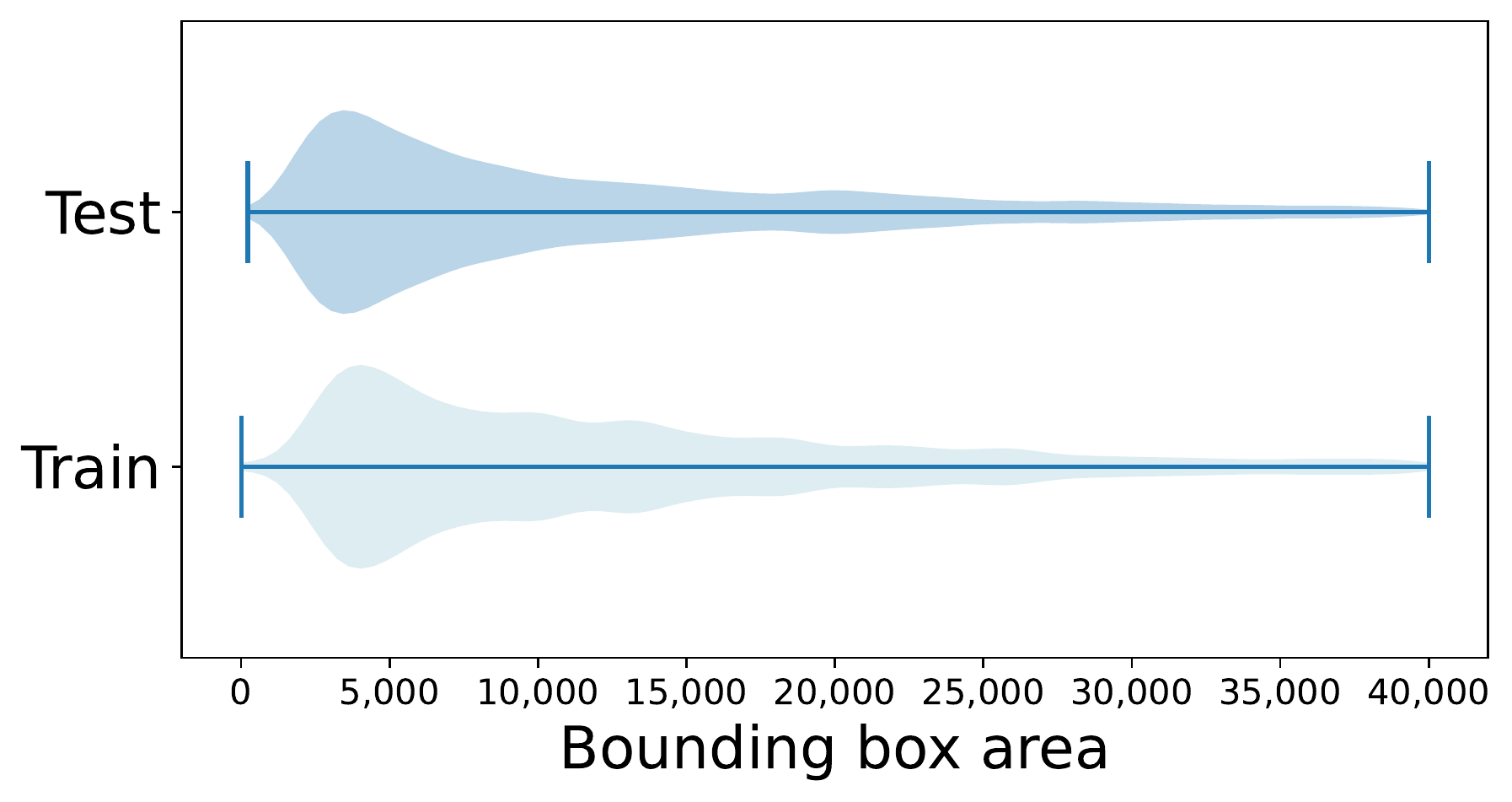}
    \caption{
    % \hrtc{TODO: Increase both axis font size} 
    Bounding box distribution of the train/validation and test splits. JRDB-Pose contains a wide distribution of pose scales. }
    \label{fig:bbox_train_test}
    \vspace{-1.5em}
\end{figure}

We started our annotation process using the JRDB bounding boxes and tracking IDs for each person in the scene. Annotators used an in-house annotation tool to manually label human poses, and all the annotations were carefully quality-assessed  multiple times by several annotators to ensure high-quality and consistent annotations. We annotated all persons from the 5 camera views of the 360º cylindrical video stream that had either a large enough bounding box (area above 6000 pixels) or clear keypoint locations. Note that this means not all visible people are labeled, especially those far away from the robot. JRDB-Pose labels poses for a wide range of scales, as shown in Fig.~\ref{fig:bbox_train_test}. We visualize the pose scales for each training and validation scene in Fig.~\ref{fig:scale_by_scene}, showing that the distribution of scales also varies significantly by scene as the types human motions and activities vary. Details on testing scenes are not shown because test data is held out.
% \hrtc{I think you might need to explain more the annotation process (ask VinAI). For example, have you done all the process manually, or initiated by an algorithm, if the later one which model and which data is trained on. then is it manually corrected? how many times and people did QA  on data? somehow ensure them Poses are very high quality  }.  

For researchers interested in panoramic views, we merge the annotations from the 5 camera views into a single 360º panoramic image following the merging protocol used in JRDB \cite{martin2021jrdb}. If a joint is labeled in two views, we use the label from the joint with a higher visibility score; if the joints have identical visibility scores, we use the pose with higher overall visibility. Poses were annotated at 7.5 Hz and finally upsampled to 15 Hz, \eddie{providing accurate high-frequency annotations without significant jitter.}
% \hrtc{Highlight again how challenging is such a frame rate annotation and we are the only one provide such temporal resolution}. 

JRDB \cite{martin2021jrdb} and JRDB-Act \cite{ehsanpour2022jrdb} previously introduced annotations including 2D and 3D bounding boxes with tracking IDs, atomic action labels, and social groups.  Together with JRDB and JRDB-Act annotations, JRDB-Pose is a multi-task learning dataset for human detection/tracking, pose estimation/tracking, individual action, social group, and social activity detection. \cref{tab:jrdb_overview} summarizes all annotations now available on videos from JRDB.

% \input{figures/5_Fig_KPTsDistributionPerSplit}

%\subsection{Occlusion}
\noindent
\textbf{Occlusion:} Occlusion is a key problem for human pose estimation and tracking because it causes incomplete information for predicting keypoints, leading to major errors in pose estimation tasks \cite{newell2016stacked}. Significant research in the field of pose estimation has focused on occlusion-aware methods for improving pose estimation \cite{achilles2016patient, cheng2019occlusion, cheng20203d_occlusion}, but may be limited by lack of quantity and diversity in the existing data \cite{cheng20203d_occlusion}. JRDB-Pose advances this research by providing large-scale data including occluded scenarios, as well as detailed occlusion labels for every keypoint which we hope will be useful for quantifying and improving performance under occlusion. In particular, we assign each keypoint a label in $\{0, 1, 2\}$ defining its occlusion type, \eddie{where for every pose we annotate the posistion and occlusion of each keypoint:}
% \begin{tabular}{||c|c|c||}

% \vspace{0.5ex}

% \renewcommand\tabularxcolumn[1]{m{#1}} % for vertical centering text
% \begin{tabularx}{\linewidth}{||c|c|X||}
%     \hline
%     ID & Meaning & Description  \\ [0.5ex]  \hline\hline
%     0 & Invisible & The joint is out of frame, especially difficult to annotate, or impossible to infer from context. \\ \hline
%     1 & Occluded & The joint is obscured by an object or another body part, but its location is apparent from the image context. \\ \hline
%     2 & Visible & The joint is fully visible and in view of the camera. \\
%     \hline
% \end{tabularx}

% \vspace{0.5ex}

% \begin{table}[!htp]\centering
% \begin{tabular}{|l|c|c|c}\toprule
% \textbf{ID} &\textbf{Meaning} &\textbf{Description} \\\cmidrule{1-3}
% \multirow{3}{*}{0} &\multirow{3}{*}{Invisible} &\multirowcell{3}[0pt][l]{The joint is out of frame, especially \\difficult to annotate, \tho{heavily occluded},\\\tho{approximately inferred by annotators.}} \\
% & & \\
% & & \\\cmidrule{1-3}
% \multirow{3}{*}{1} &\multirow{3}{*}{Occluded} &\multirowcell{3}[0pt][l]{The joint is obscured by an object or\\another body part, but its location is\\apparent from the image context.} \\
% & & \\
% & & \\\cmidrule{1-3}
% \multirow{2}{*}{2} &\multirow{2}{*}{Visible} &\multirowcell{2}[0pt][l]{The joint is fully visible and in view\\of the camera.} \\
% & & \\
% \bottomrule
% \end{tabular}
% \end{table}

\begin{table}[htp]\centering
\small
\begin{tabular}{p{4mm}|p{16mm}|p{50mm}} \hline 
\textbf{ID} &\textbf{Meaning} &\textbf{Description} \\ [0.5ex] \hline
0 & Invisible & The joint is out of frame or especially difficult to annotate, \eddie{with a location inferred from context by annotators}. \\ \hline
1 & Occluded & The joint is obscured by an object or another body part, but its location is apparent from the image context. \\ \hline
2 & Visible & The joint is fully visible and in view of the camera. \\
\hline

\end{tabular}
\vspace{-1em}
\end{table}

% \end{tabular}

\begin{figure}
    \centering
    \includegraphics[width=\linewidth]{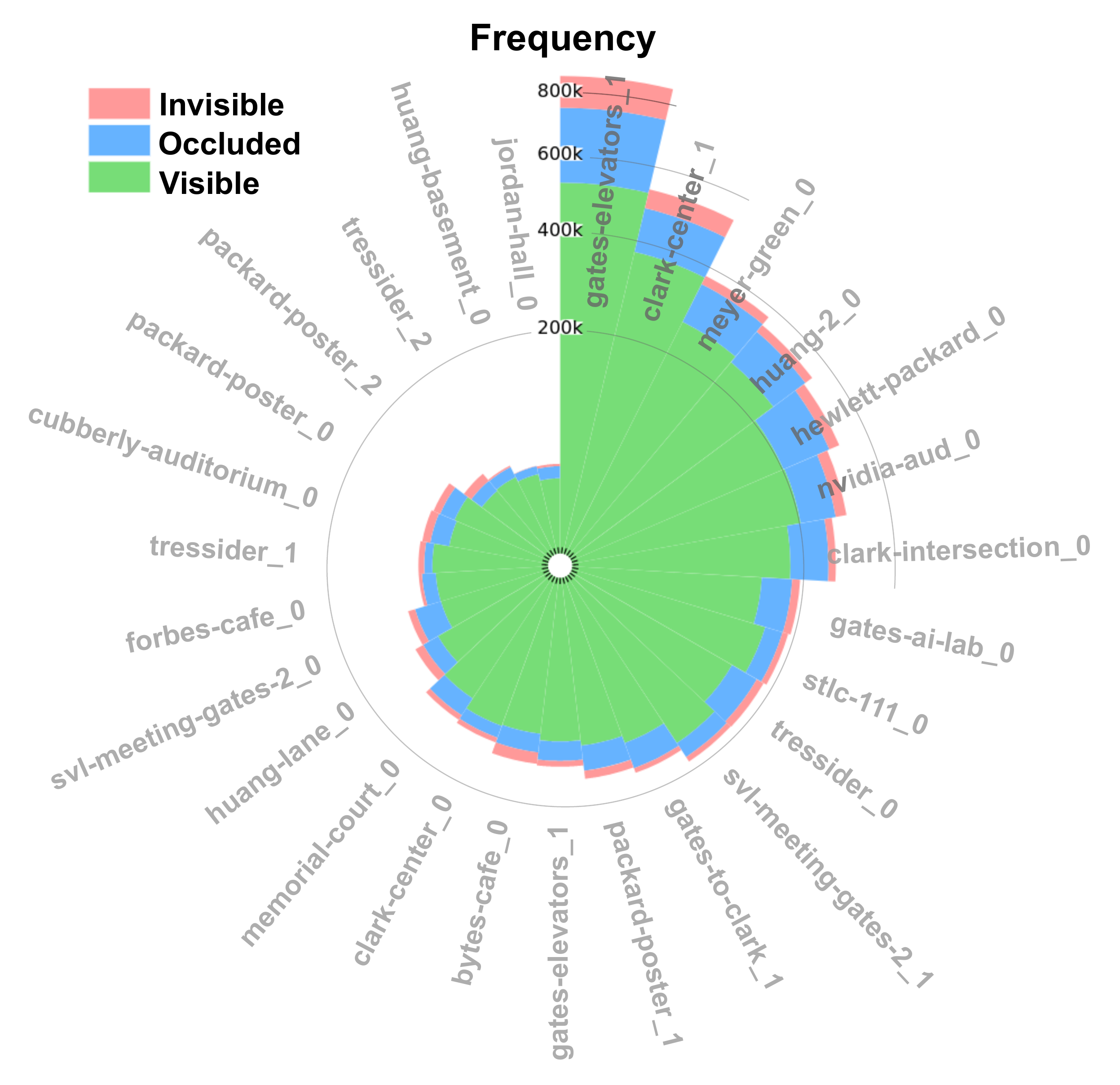}\vspace{-.5em}
    %  \vspace{-0.5em}
    \caption{Distribution of keypoint visibility annotations for each JRDB-Pose train and validation scene. Best viewed in color.}
    \vspace{-1em}
    \label{fig:vis_by_scene}
\end{figure}

Figure \ref{fig:vis_train_test} shows the occlusion label distribution in the train and test splits. JRDB-Pose contains a large number of labeled occluded and invisible joints, making it suitable for pose estimation under occlusion. Occlusion varies by scene: \cref{fig:vis_by_scene} shows that the training scenes of JRDB-Pose contain varied distributions of keypoint visibility, from densely populated indoor to dynamic outdoor scenes in which the robot navigates and interacts with oncoming pedestrians.

\subsection{Benchmark and Metrics}

\textbf{JRDB-Pose Splits:}
We follow the splits of JRDB \cite{martin2021jrdb} to create training, validation, and testing splits from the 54 captured sequences, with each split containing an equal proportion of indoor and outdoor scenes as well as scenes captured using a stationary or moving robot. All frames from a scene appear strictly in one split.

\subsubsection{Pose estimation}

\noindent
\textbf{OKS}. We define the ground-truth and a corresponding predicted human body keypoints as $x_i\in \mathbb{R}^{2\times J}$ and $y_j \in \mathbb{R}^{2\times J}$, respectively, where $J$ represents the number of the keypoints. We measure a similarity score between $x_i$ and $y_j$ via Object Keypoint Similarity (OKS)~\cite{lin2014microsoftcoco} as:
\begin{equation}
\label{eq:OKS}
O K S(x_i,y_j)=\exp \left(-\frac{d_E^2(x_i,y_j)}{2 s^2 k^2}\right)
\end{equation}
where $d_E(x_i,y_j)$ is the mean Euclidean distance between two set of keypoints normalized by the product of ground-truth box area $s$ and a sigma constants $k$. While this metric is commonly used for single-person pose estimation, in multi-person settings the assignment between ground-truth and predicted poses is not known, so it is unclear which pose pairs to use without a matching mechanism.\\
\noindent
\textbf{Average Precision}. Average Precision (AP) and mean AP (mAP) \cite{pishchulin2016deepcut} are among the most common multi-person pose estimation metrics. Similarity between predicted and ground-truth poses are calculated via OKS\cite{lin2014microsoftcoco}, and poses are matched via a non-optimal greedy assignment strategy that matches poses with higher confidence scores first. True and false positive predictions are determined by a threshold on the OKS score. Since the resulting AP score corresponds to the performance at a specific OKS threshold rather than a continuous value based on overall keypoint distances, the COCO benchmark \cite{lin2014microsoftcoco} averages AP across multiple OKS thresholds to reward methods with higher localization accuracy. Nevertheless, this strategy does not fully capture keypoint distance nor is the matching strategy optimal. We evaluate AP using an OKS threshold of $0.5$.

\noindent
\textbf{OSPA-Pose}. Optimal Sub-Pattern Matching (OSPA) \cite{schuhmacher2008consistent} is a multi-object performance evaluation metric which includes the concept of miss-distance. Recently, OSPA has been further adapted to detection and tracking tasks \cite{rezatofighi2020trustworthy} while preserving its metric properties. We propose OSPA-Pose ($\mathcal{O}_{pose}$), a modification of the OSPA metric for multi-person pose estimation.

Let $X = \{x_1, x_2, \dots x_m\}$ and $Y = \{y_1, y_2, \dots y_n\}$ be two sets of body keypoints for all ground-truth and predicted body poses, respectively. Considering $d_{K}(x_i, y_i) = 1 - OKS(x_i, y_i) \in [0, 1]$ as a normalized distance metric between human poses, OSPA-Pose ($\mathcal{O}_{pose}$) error is calculated by
\begin{equation}
    \mathcal{O}_{pose}(X, Y) \!= \!\frac{1}{n}\bigg(\!\!
\min_{\pi \in \Pi_n}\!\!
\sum_{i=1}^m (d_K(x_i, y_{\pi_i}))\! +\! (n-m)\!\!
\bigg),
\label{eq:ospa}
\end{equation}
where $n \geq m \ge 0$, $\Pi_n$ is the set of all permutations of $\{1, 2, \dots, n\}$, and $\mathcal{O}_{pose}(X, Y)$ = $\mathcal{O}_{pose}(Y, X)$ if $m < n$. We further define $\mathcal{O}_{pose}(X, Y) = 1$ if either $X$ or $Y$ is empty, and $\mathcal{O}_{pose}(\emptyset, \emptyset) = 0$.  

While both AP and $\mathcal{O}_{pose}$ use OKS to calculate pose similarity, $\mathcal{O}_{pose}$ measures an optimal distance from the set of continuous keypoint distances consisting of the localization error (first term) and cardinality mismatch (second term), eliminating the need for thresholding. %The first term in $\mathcal{O}_{pose}$ represents  

% \begin{table*}[t]
%     \centering
%     \begin{tabular}{cccccccccc}
%         \toprule
%               &  &  &  &\multicolumn{3}{c}{Total} & \multicolumn{3}{c}{Per-Keypoint} \\  \cmidrule(lr){5-7} \cmidrule(lr){8-10}
% Method      & OSPA-Pose & Localization  & Cardinality & V     & O & I & V & O & I &  \\  [0.5ex] \hline 
% % HRNet W48   & 0.459     & \xxx & \xxx & \xxx  & \xxx  & \xxx  & \xxx  & \xxx  & \xxx  &      42.34 & \xxx \\
% % ResNet-50   &     0.450 & \xxx & \xxx & \xxx  & \xxx  & \xxx  & \xxx  & \xxx  & \xxx  &      43.34 & \xxx \\
% % Yolo-Pose   &     0.450 & \xxx & \xxx & \xxx  & \xxx  & \xxx  & \xxx  & \xxx  & \xxx  &      43.34 & \xxx \\
% HRNet W48 & 48.14 & 23.48 & 24.66 & 14.69 & 6.03 & 2.76 & 0.14 & 0.2 & 0.26 \\
% DEKR & 0 & 0 & 0 & 0 & 0 & 0 & 0 & 0 & 0 \\
% CID & 37.74 & 17.38 & 20.35 & 10.11 & 4.71 & 2.57 & 3.82 & 6.33 & 10.8 \\
% YoloPose & 35.72 & 0 & 0 & 0 & 0 & 0 & 0 & 0 & 0 \\
%         \bottomrule
%     \end{tabular}
%     \caption{We measure the performance of our baseline multi-person pose estimation models with OSPA-Pose. Loc and Card are the OSA localization and cardinality error, respectively. We measure performance by joint visibility via the contribution to the OSPA-Pose score from joints of different visibilities. A lower score indicates that the model performs better. }
%     \label{tab:results}
% \end{table*}

\subsubsection{Pose tracking metrics}

\textbf{Commonly used metrics}: We evaluate pose tracking performance with three commonly used tracking metrics, \textbf{MOTA} \cite{andriluka2018posetrack}, \textbf{IDF1} \cite{ristani2016performance}, and \textbf{IDSW} \cite{milan2016mot16}, with modifications to make them suitable for the pose tracking task. Rather than using IoU or GIoU\cite{rezatofighi2019generalized} to calculate similarity scores which are more appropriate for bounding boxes, we apply OKS (as defined in \cref{eq:OKS}) to obtain a similarity score between keypoint sets, with any keypoint pair of OKS above a threshold of 0.5 considered a positive prediction.
\\
\textbf{OSPA$^{(2)}$-Pose} Inspired by~\cite{beard2020solution, rezatofighi2020trustworthy} extending the OSPA metric for evaluating two sets of trajectories, we propose a new metric for evaluating two sets of human body pose tracks, namely OSPA$^{(2)}$-Pose (\ospatrack). 

Let $\mathbf{X} = \{X^{\mathcal{D}_1}_1, X^{\mathcal{D}_2}_2, \dots X^{\mathcal{D}_m}_m\}$ and $\mathbf{Y} = \{Y^{\mathcal{D}_1}_1, Y^{\mathcal{D}_2}_2, \dots Y^{\mathcal{D}_n}_n\}$ to be two sets of body keypoint trajectories for ground-truth and predicted body poses, respectively. Note $\mathcal{D}_i$ represents the time indices which track $i$ exists (having a state-value). Then, we calculate the time average distance of every pair of tracks $X^{\mathcal{D}_i}_i$ and $Y^{\mathcal{D}_j}_j$:
\begin{equation}
\label{eq:time-av}
\underline{\widetilde{d}}(X^{\mathcal{D}_i}_i, Y^{\mathcal{D}_j}_j)
=\sum_{t \in \mathcal{D}_i \cup \mathcal{D}_j} \frac{d_{O}\left(\{X_i^{t}\},
\{Y_j^{t}\}\right)}{|\mathcal{D}_{i} \cup \mathcal{D}_{j}|},
\end{equation}
where $t \in \mathcal{D}_{i} \cup \mathcal{D}_{j}$ is the time-step when either or both track presents. Note that $\{X_i^{t}\}$ and $\{Y_j^{t}\}$ are singleton sets, \ie $\{X_i^{t}\} = \emptyset$ or $\{X_i^{t}\} = x_i^t\in \mathbb{R}^{2\times J}$ and $\{Y_j^{t}\} = \emptyset$ or $\{Y_j^{t}\} = y_j^t\in \mathbb{R}^{2\times J}$. Therefore, inspired by the OSPA set distance, $d_{O}\left(\{X_i^{t}\} , \{Y_j^{t}\}\right)$ can be simplified into the following distance function, :
\begin{align}
d_{O}(\{X_i^{t}\},&\{Y_j^{t}\})= \nonumber\\
&\left\{\begin{array}{lll}
d_K(x^t_i, y^t_j) &\text{if } |\{X_i^{t}\}|\wedge|\{Y_j^{t}\}|=1, \\
1 &\text{if } |\{X_i^{t}\}|\oplus|\{Y_j^{t}\}|=1,\\
0 &\text{Otherwise,}
\end{array}\right.
\end{align}
where $d_{K}(x_i, y_i) = 1 - OKS(x_i, y_i) \in [0, 1]$.

Finally, we obtain the distance, \ospatrack, between two sets of pose tracks, \ie $\mathbf{X}$ and $\mathbf{Y}$ by applying another OSPA distance over \cref{eq:time-av}, \ie 
\begin{align}
 \mathcal{O}^2_{pose}&(\mathbf{X},\mathbf{Y})=\nonumber\\ \frac{1}{n}&\left(\min_{\pi \in \Pi_{n}} \sum_{i=1}^{m}\underline{\widetilde{d}}(X^{\mathcal{D}_i}_i, Y^{\mathcal{D}_{\pi_i}}_{\pi_i})+(n-m)\right),
\end{align}
where $n \geq m \ge 0$, $\Pi_n$ is the set of all permutations of $\{1, 2, \dots, n\}$, and $\mathcal{O}^2_{pose}(X, Y)$ = $\mathcal{O}^2_{pose}(Y, X)$ if $m < n$. We further define $\mathcal{O}^2_{pose}(X, Y) = 1$ if either $X$ or $Y$ is empty, and $\mathcal{O}^2_{pose}(\emptyset, \emptyset) = 0$. 

\tho{Note that the first term $\underline{\widetilde{d}}$ reflects ID switches and localization errors, whereas the cardinality error $(n-m)$ contains false and missed track errors. In \cref{tab:results_tracking}, we also present the \ospatrack \ per occlusion level, where each occlusion level is considered individually. We do not apply localisation error to joints that do not belong to the occlusion of interest by setting their Euclidean distance to zero, indicating that $d_E(x_i,y_j)$ in \cref{eq:OKS} is now the average Euclidean distance between two sets of keypoints of a certain occlusion level. }

\section{Multi-Person Pose Estimation and Tracking Baselines}

In this section we evaluate the performance of various state-of-the-art methods for pose estimation and tracking, verifying that JRDB-Pose is a challenging dataset for existing frameworks seeking an opportunity for a dedicated developments in this domain. All methods are evaluated on our individual images and annotations. 
% \hrtc{This is not correct, we report on indvidual camera}

\subsection{Multi-Person Pose Estimation}
We evaluate several popular or recent state-of-the-art methods for multi-person pose estimation models \cite{wang2020deep, geng2021bottom, wang2022contextualcid, maji2022yolo}. We evaluate one top-down method using HRNet backbone \cite{wang2020deep}. The top-down method uses a Faster R-CNN \cite{ren2015fasterrcnn} detector to predict all humans in the scene, from which the high-confidence (score $>$ 0.95) predictions are kept. We use a heatmap size of 192x256, and the HRNet backbone with a high-resolution convolution width of 48. We further evaluate three recent bottom-up models, which regress joint locations directly without the need for human detections: DEKR \cite{geng2021bottom}, CID \cite{wang2022contextualcid}, an YoloPose \cite{maji2022yolo}. All methods are trained from their respective initializations without COCO pre-training. To help address training difficulties associated with wide panoramic images, we train on individual camera images, and then combine them together to form stitched view predictions. Duplicate poses in the stitched annotation set are eliminated using NMS on the predicted boxes.

% \begin{table}[t]
%     \centering
%     \begin{tabular}{c|ccccc}
%         \toprule
% Method      & \ospa$\downarrow$ & Loc  & Card & AP$^{0.5}$ & AR  \\  [0.5ex] \hline 
% HRNet \cite{wang2020deep} & 48.0 & 21.0 & 27.0 & 34.60 & 87.34 \\ 
% DEKR \cite{geng2021bottom}& \xxx & \xxx & \xxx & \xxx & \xxx \\
% CID~\cite{wang2022contextualcid} & 37.7 & 17.4 & 20.4 & \xxx & \xxx \\
% YOLO-Pose\cite{maji2022yolo} & \textbf{35.7} & \xxx & \xxx & \xxx & \xxx \\
%         \bottomrule
%     \end{tabular}
%     \caption{Multi-person pose estimation baselines evaluated on JRDB-Pose. Loc and Card are the OSPA localization and cardinality error, respectively. AP$^{OKS=k}$ denotes AP with a positive detection threshold of keypoint OKS above $k$.}
%     \label{tab:results_summary}
% \end{table}

\begin{table}[!tb]\centering
\footnotesize
\resizebox{\linewidth}{!}{%
\begin{tabular}{c|cccccc}\toprule
\textbf{Method} &\ospa &Loc &Card &AP &AR \\\cmidrule{1-6}
HRNet\cite{wang2020deep} &0.480 &0.210 &0.270 & 24.6  & 47.5 \\
DEKR\cite{geng2021bottom} &0.410 &\textbf{0.113} & 0.2968 & 31.7 & 46.9 & \\
CID\cite{wang2022contextualcid} &0.377 &0.174 &0.204 &38.6 & 48.0 \\
YOLO-Pose\cite{maji2022yolo} &\textbf{0.368} &0.172 &\textbf{0.196} & \textbf{47.9} & \textbf{72.5} \\ \cmidrule{1-6}
\end{tabular}}\vspace{-.5em}
    \caption{Multi-person pose estimation baselines evaluated on JRDB-Pose stitched annotations. Loc and Card are the \ospa localization and cardinality error, respectively.} \vspace{-.5em}
    \label{tab:results_summary}
\end{table}

% \begin{table}[t]
%     \centering
%     \begin{tabular}{ccccccc}
%         \toprule
%               & \multicolumn{3}{c}{Total Contribution} & \multicolumn{3}{c}{Per-Keypoint Avg} \\  \cmidrule(lr){2-4} \cmidrule(lr){5-7}
% Method      & Vis   & Occ   & Inv   & Vis   & Occ   & Inv \\  [0.5ex] \hline 
% HRNet~\cite{wang2020deep} & 13.1 & 5.4 & 2.5 & 5.1 & 7.5 & 10.2 \\
% DEKR~\cite{geng2021bottom} & \xxx & \xxx & \xxx & \xxx & \xxx & \xxx \\
% CID~\cite{wang2022contextualcid} & 10.1 & 4.7 & 2.6 & 3.8 & 6.3 & 10.8 \\
% YOLO-Pose~\cite{maji2022yolo} & \xxx & \xxx & \xxx & \xxx & \xxx & \xxx \\
%         \bottomrule
%     \end{tabular}
%     \caption{We measure performance by joint visibility via the contribution to the OSPA-Pose score from joints of different visibilities. A lower score indicates that the model performs better.}
%     \label{tab:results}
% \end{table}

\begin{table}[!tb]\centering
\small
\resizebox{\linewidth}{!}{%
\begin{tabular}{c|ccc|ccc}\toprule
\multirow{2}{*}{\textbf{Method}} &\multicolumn{3}{c}{Total Contribution} &\multicolumn{3}{c}{Per-Keypoint Avg} \\
&V$\downarrow$ &O$\downarrow$ &I$\downarrow$ &V$\downarrow$ &O$\downarrow$ &I$\downarrow$ \\\cmidrule{1-7}
HRNet\cite{wang2020deep} &0.131 &0.054 &0.025 &0.051 &0.075 &0.102 \\
DEKR\cite{geng2021bottom} & \textbf{0.070} &\textbf{0.031 }& \textbf{0.012 }& \textbf{0.030} & \textbf{0.052} & \textbf{0.089} \\
CID\cite{wang2022contextualcid} &0.101 &0.047 &0.026 &0.038 &0.063 &0.108 \\
YOLO-Pose\cite{maji2022yolo} &0.098 &0.048 &0.026 &0.033 &0.055 &0.091 \\\cmidrule{1-7}
\end{tabular}} \vspace{-.5em}
\caption{We break down \ospa localization error from \cref{tab:results_summary} \eddie{ for visible (V), occluded (O), and invisible (I) joints. The left and right columns represent the total and average per-keypoint contribution (scaled by 1M) to the \ospa localization error, respectively.}} \vspace{-.5em}
\label{tab:results}
\vspace{-1em}
\end{table}

\begin{table*}[!tb]\centering
\footnotesize
\resizebox{\linewidth}{!}{%
\begin{tabular}{cccccccccccc}\toprule
\multirowcell{2}{\textbf{Pose Estimation} \\ \textbf{Method} {\scriptsize(Training)}} &\multirowcell{2}{\textbf{Tracking} \\ \textbf{Method}} &\multirow{2}{*}{MOTA $\uparrow$} &\multirow{2}{*}{IDF1$\uparrow$} &\multirow{2}{*}{IDSW$\downarrow$} &\multirow{2}{*}{\ospatrack$\downarrow$} &\multicolumn{2}{c}{Components} &\multicolumn{3}{c}{\ospatrack$\downarrow$ by Visibility} \\\cmidrule{7-11}
& & & & & &Card$\downarrow$ &Loc$\downarrow$ &V$\downarrow$ &O$\downarrow$ &I$\downarrow$ \\\cmidrule{1-11}
\multirowcell{3}{Yolo-Pose\cite{maji2022yolo} \\ \scriptsize(COCO only)} &ByteTrack\cite{bytetrack} &61.32 &55.80 &4236 &0.715 &0.519 &\textbf{0.196} &0.698 &0.703 &0.731 \\
&UniTrack\cite{unitrack} &60.80 &55.01 &2854 &0.727 &0.473 &0.253 &0.713 &0.716 &0.733 \\
&OCTrack\cite{ocsort} &\textbf{66.09} &\textbf{59.93} &\textbf{2588} &\textbf{0.630} &\textbf{0.380} &0.249 &\textbf{0.610} &\textbf{0.618} &\textbf{0.646} \\\cmidrule{1-11}
\multirowcell{3}{Yolo-Pose\cite{maji2022yolo} \\ \scriptsize(JRDB-Pose only)}&ByteTrack\cite{bytetrack} &61.17 &52.11 &3203 &0.693 &0.429 &\textbf{0.264} &0.678 &0.692 &0.708 \\
&UniTrack\cite{unitrack} &\textbf{69.84} &56.06 &2565 &0.725 &0.454 &0.271 &0.710 &0.722 &0.734 \\
&OCTrack\cite{ocsort} &69.22 &\textbf{60.62} &\textbf{1977} &\textbf{0.577} &\textbf{0.295} &0.283 &\textbf{0.556} &\textbf{0.577} &\textbf{0.595} \\\cmidrule{1-11}
\multirowcell{3}{Yolo-Pose\cite{maji2022yolo} \\ \scriptsize(COCO$\rightarrow$ \\ \scriptsize  JRDB-Pose)} &ByteTrack\cite{bytetrack} &67.16 &55.38 &3325 &0.690 &0.456 &\textbf{0.234} &0.674 &0.688 &0.708 \\
&UniTrack\cite{unitrack} &\textbf{72.82} &57.69 &2413 &0.708 &0.435 &0.272 &0.693 &0.707 &0.719 \\
&OCTrack\cite{ocsort} &71.74 &\textbf{61.15} &\textbf{2260} &\textbf{0.594} &\textbf{0.331} &0.263 &\textbf{0.573} &\textbf{0.592} &\textbf{0.613} \\
\bottomrule
\end{tabular}
}\vspace{-.5em}
\caption{Multi-person pose tracking baselines evaluated on JRDB-Pose individual camera images.} \vspace{-1em}
\label{tab:results_tracking}

\end{table*}

\cref{tab:results_summary} summarises the pose estimation results for each method. To highlight the large number of occluded and invisible labels in our dataset, we include the total contribution of joints of each visibility type to the overall OSPA-Pose as well as average contribution for joints of each visibility. We also include cardinality and localization errors of OSPA-Pose. \eddie{We find that Yolo-Pose is the best performing baseline, outperforming the other methods in \ospa and AP. DEKR achieves the lowest \ospa localization error but a higher cardinality error as a result of a high number of missed detections. In \cref{tab:results} we see the contributions to the localization error based on keypoint visibility. DEKR achieves the lower metrics due to its low localization error, besides which Yolo-Pose again performs best. Although visible joints contribute the most to localization error due to their higher overall frequency, on a per-keypoint average the predictions on occluded and invisible joints showed significantly higher errors for all models, confirming that occlusion poses a difficult challenge for existing methods. JRDB-Pose contains a significant number of occluded joints which we hope will be useful for researchers to improve the robustness of pose estimation methods to occlusions. We also find that all methods achieve lower AP as compared to their results on common large-scale benchmarks \cite{lin2014microsoftcoco}, showing that JRDB-Pose presents a reasonable and difficult challenge for multi-person pose estimation.}

\subsection{Multi-Person Pose Tracking}

We evaluate three recent state-of-the-art methods, such as  ByteTrack \cite{bytetrack}, Unitrack \cite{unitrack}, and OC-SORT \cite{ocsort}, in our tracking benchmark. All these methods are from the tracking-by-detection category and predict tracks based on predictions from a pose estimation method. We use the estimates from Yolo-Pose, as our highest-performing baseline, to initialize the predictions for all pose tracking methods.

In \cref{tab:results_tracking} we provide our pose tracking results for all trackers and training strategies. OC-Sort achieves the best results by a wide margin. Since OC-SORT targets improving tracking robustness for objects in non-linear motion with the improved Kalman-filter and recovery strategy, we believe it is better suited for JRDB-Pose which includes occluded periods during which people are often not moving linearly with respect to the robot's perspective \tho{(\eg sequences where both the robot and nearby people are turning or moving simultaneously)}, thus better recovering from occlusions. This method's lower \ospatrack  cardinality error further confirms that it can better find accurate tracks. Compared to their performance in other large-scale tracking benchmarks~\cite{doering2022posetrack21, dendorfer2020mot20}, these methods achieve relatively lower overall performance in the same reported metrics, reflecting the unique challenges in our dataset, which we hope will motivate further work within the research community. 

\subsection{Study on Pre-training Methods}
Using our highest performing pose estimation method, we further study how pre-training strategies affect our final model performance. We try inference-only with a COCO-pretrained model, finetuning a COCO-pretrained model, and training from scratch. \cref{tab:yolopretraining} and \cref{tab:results_tracking} show pose estimation and tracking performance of our highest performing model across the different training strategies. For pose estimation we find that finetuning from COCO generally performs better than training from scratch in both pose estimation and tracking tasks with an improvement of $1.7\%$ in \ospatrack, while running inference from a COCO model without finetune shows much weaker results especially in \ospa. The relatively small improvement also suggests that JRDB-Pose contains a varied distribution of scenes and poses suitable for model training.

\eddie{We also include results for pose tracking across the same three training strategies of Yolo-Pose used to initialize the tracking. In \cref{tab:results_tracking} we find that the JRDB-Pose model outperforms the COCO model finetuned on JRDB-Pose in MOTA and IDF1 but is itself outperformed in IDSW and \ospatrack. The COCO model without finetuning performs worse than the other models. Since the COCO model is not trained on the crowded human scenes and overall image distribution present in JRDB-Pose, it struggles to find keypoints and thus has a high miss-rate as indicated by a much higher \ospatrack cardinality component compared to the other training schemes. JRDB-Pose contains a varied distribution of scenes in human environments different from COCO.}

% \begin{table}[t]
%     \centering
%     \begin{tabular}{ccccc}
%         \toprule
%       &  \multicolumn{2}{c}{Stitched}  & \multicolumn{2}{c}{Individual Camera} \\  \cmidrule(lr){2-3} \cmidrule(lr){4-5}
% Training & AP & \ospa & AP & \ospa  \\  [0.5ex] \hline 
% COCO & 48.0              & 46.0          & 60.2 & 32.0 \\
% JRDB-Pose & 48.2           & \textbf{35.7} & 61.9 & 28.0 \\
% Finetune & \textbf{49.6} & 38.5          & \textbf{63.9} & \textbf{27.3} \\
%         \bottomrule
%     \end{tabular}
%     \caption{We study how training protocols affect performance of YoloPose. Finetuning from COCO achieves best results.}
%     \label{tab:yolopretraining}
% \end{table}

\begin{table}[!tb]\centering
\small
\resizebox{\linewidth}{!}{%
\begin{tabular}{c|ccccc}\toprule
\multirowcell{2}{\textbf{Training} \\ \textbf{strategy}} &\multicolumn{2}{c}{\textbf{Stitched}} &\multicolumn{2}{c}{\textbf{Per Camera}} \\\cmidrule{2-5}
&AP $\uparrow$ &\ospa $\downarrow$ &AP $\uparrow$ &\ospa $\downarrow$ \\\cmidrule{1-5}
COCO \cite{lin2014microsoftcoco} &48.0 &0.460 &60.2 &0.320 \\
JRDB-Pose &48.2 &\textbf{0.357} &61.9 &0.280 \\
COCO $\rightarrow$ JRDB-Pose &\textbf{49.6} &0.385 &\textbf{63.9} &\textbf{0.273} \\\midrule
\end{tabular}
}\vspace{-.5em}
    \caption{We study how training protocols affect performance of YoloPose. Finetuning from COCO achieves best results.}\vspace{-2em}
    \label{tab:yolopretraining}
\end{table}

% \begin{table}[!htp]\centering
% \begin{tabular}{c|ccccc}\toprule
% \multirowcell{2}{Training \\ strategy} &\multicolumn{2}{c}{Stitched} &\multicolumn{2}{c}{Per Camera} \\\cmidrule{2-5}
% &AP $\uparrow$ &\ospa $\downarrow$ &AP $\uparrow$ &\ospa $\downarrow$\\\cmidrule{1-5}
% COCO &48.0 &46.0 &60.2 &32.0 \\\cmidrule{1-5}
% JRDB-Pose &48.2 &\textbf{35.7} &61.9 &28.0 \\\cmidrule{1-5}
% \multirowcell{2}{\footnotesize COCO \\ \footnotesize $\rightarrow$ JRDB-Pose} &\multirow{2}{*}{\textbf{49.6}} &\multirow{2}{*}{38.5} &\multirow{2}{*}{\textbf{63.9}} &\multirow{2}{*}{\textbf{27.3}} \\
% & & & & \\
% \bottomrule
% \end{tabular}
%     \caption{We study how training protocols affect performance of YoloPose. Finetuning from COCO achieves best results.}
%     \label{tab:yolopretraining}
% \end{table}

%%%%%%%%% CONCLUSION
\section{Conclusion}

In this paper we have introduced JRDB-Pose, a large-scale dataset of human poses and track IDs suitable for multi-person pose estimation and tracking from videos. JRDB-Pose features high-frequency annotations for crowded indoor and outdoor scenes with heavy occlusion, diverse actions, and longer videos than existing large-scale pose estimation and tracking datasets. Finally, we have introduced OSPA-Pose and OSPA$^(2)$-Pose, new metrics for multi-person pose estimation and tracking. We believe that JRDB-Pose will help address limitations in human action understanding for human-robot interaction and navigation in human environments and advance research by proving large-scale and high-quality annotations.

%%%%%%%%% REFERENCES
{\small
\bibliographystyle{ieee_fullname}
\bibliography{egbib}
}

\pagebreak

\appendix

% \begin{center}
\noindent
% \textbf{\large Supplemental Materials for JRDB-Pose: A Large-scale Dataset for Multi-Person Pose Estimation and Tracking}
\textbf{\Large Appendix}

% \end{center}

\section{Annotation Visualization}
\begin{figure}
    \centering
    \includegraphics[width=\linewidth]{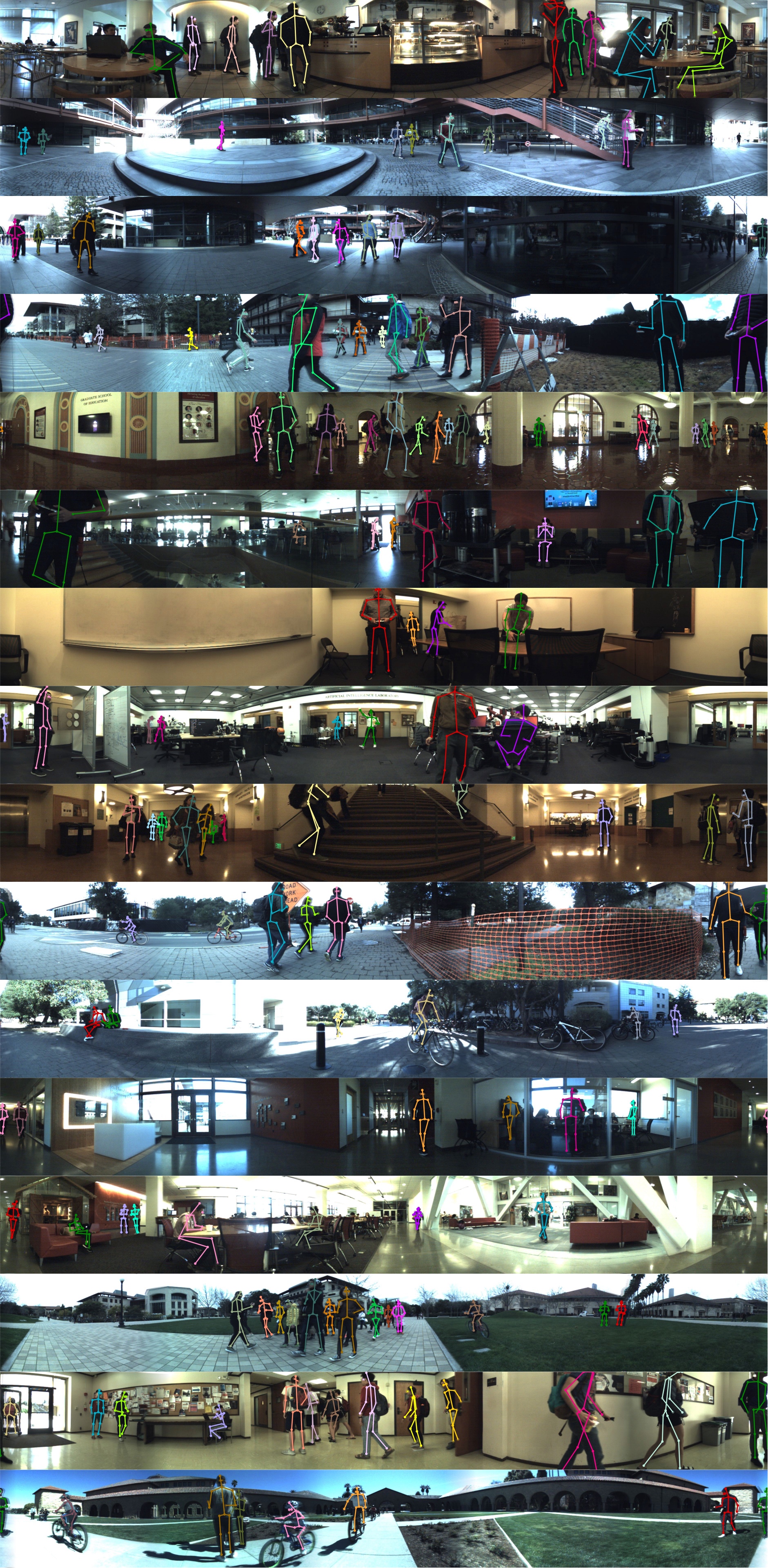}
    \vspace{-.5em}
    %  \vspace{-0.5em}
    \caption{Visualization of JRDB-Pose annotations from each of the 27 training sequences in JRDB-Pose. The images feature indoor and outdoor areas on a university campus with varying lighting conditions, motion, pedestrian density, and activities. Locations include roads, strip malls, sidewalks, restaurants, plazas, parks, halls, classrooms, laboratories, and office buildings. These scenes show a wide range of scenarios and human poses that a social robot would encounter during operations. }
    \vspace{-1em}
    \label{fig:trainsetviz16}
\end{figure}

\begin{figure}[!t]
    \centering
    \vspace{-3em}
    \includegraphics[width=\linewidth]{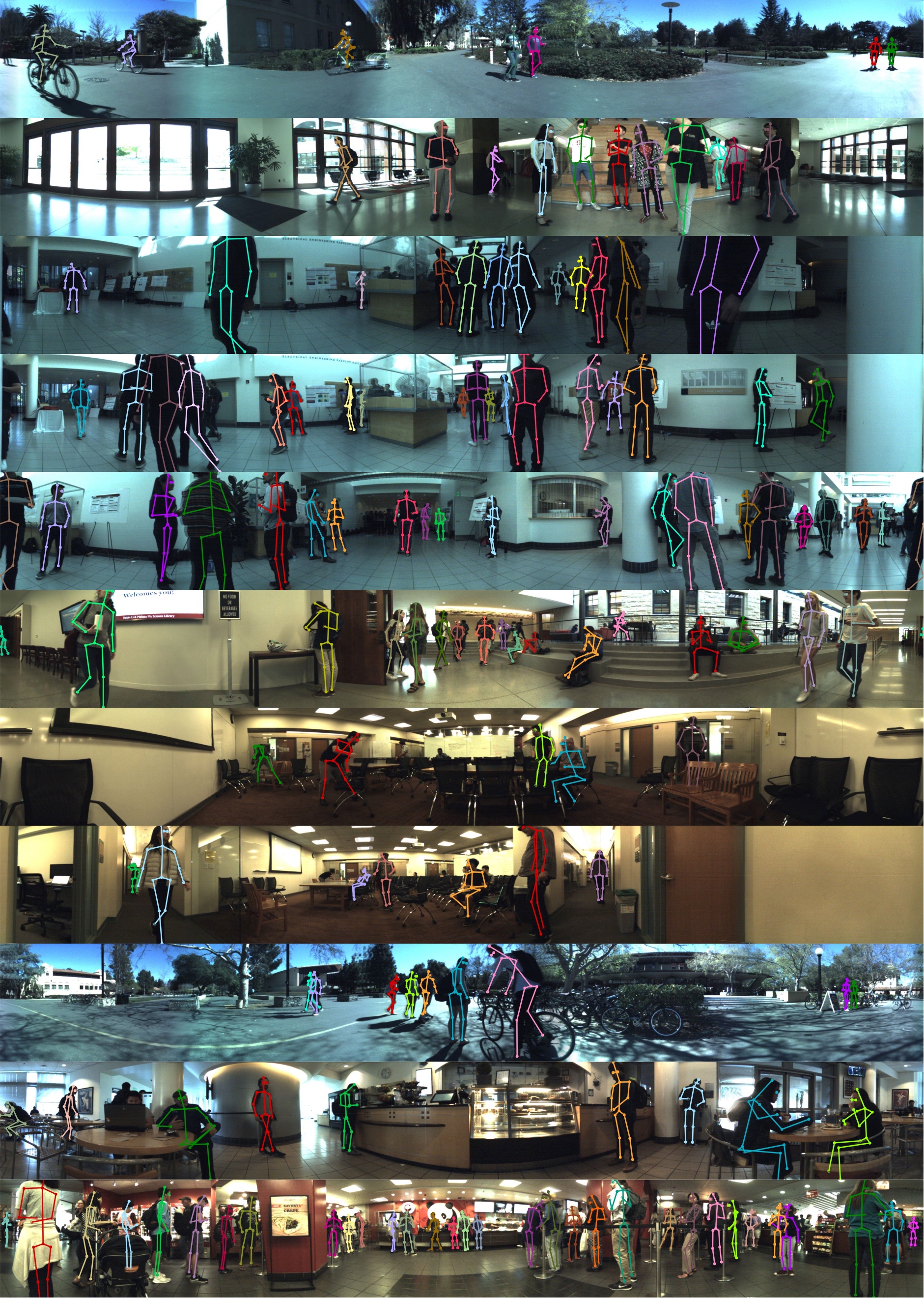}
    \vspace{-.5em}
    \caption{Visualization of JRDB-Pose annotations on the 27 training sequences (cont.) }
    \vspace{-1em}
    \label{fig:trainsetviz11}
\end{figure}

JRDB-Pose provides high-quality pose annotations at a high (15fps) frequency over all JRDB scenes. In Figure \cref{fig:trainsetviz11} and Figure \cref{fig:trainsetviz16} we visualize of our pose annotations for each of the training scenes. The visualizations show the indoor and outdoor scenes, varying light conditions, and a diverse poses representing a wide distribution of actions. To highlight the high frequency and quality of annotations we also provide a video in the supplementary material titled \textbf{JRDB\_Pose\_Supp\_Video.mp4}. 

\section{Additional Evaluations}

In addition to the evaluation we provide in the main paper, we also include metrics for additional modalities and camera types. In \cref{tab:results_tracking_stitched} we report the tracking results for our baseline models and pre-training methods using the panoramic stitched camera images. Similar to the results on individual camera images, OCTrack outperforms the other baselines. We also similarly observe that fine-tuning from COCO yields mixed results with improvements in MOTA and IDF1 but drops in the other metrics as compared to training on JRDB-Pose only.

\section{Qualitative Analysis}

\begin{figure}[!t]
    \centering
    \includegraphics[width=\linewidth]{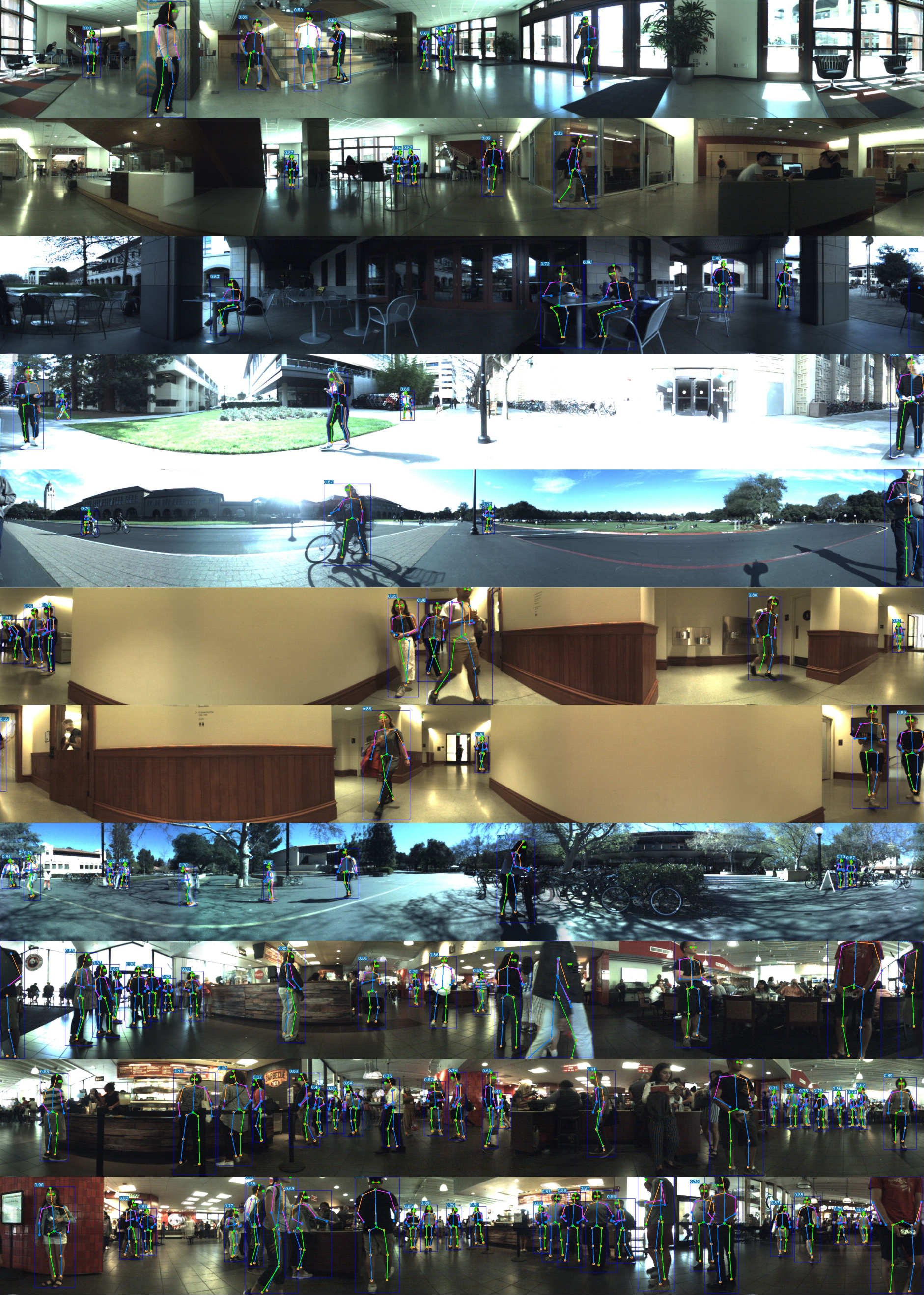}
    \vspace{-.5em}
    %  \vspace{-0.5em}
    \caption{Visualization of Yolo-Pose pose estimation predictions on 27 testing sequences in JRDB-Pose. The weights of the model are initialized from scratch, demonstrating that JRDB-Pose is sufficiently large for the model to learn accurate pose representations even without finetuning.}
    \vspace{-1em}
    \label{fig:testsetviz11}
\end{figure}

\begin{figure}[!t]
    \centering
    \includegraphics[width=\linewidth]{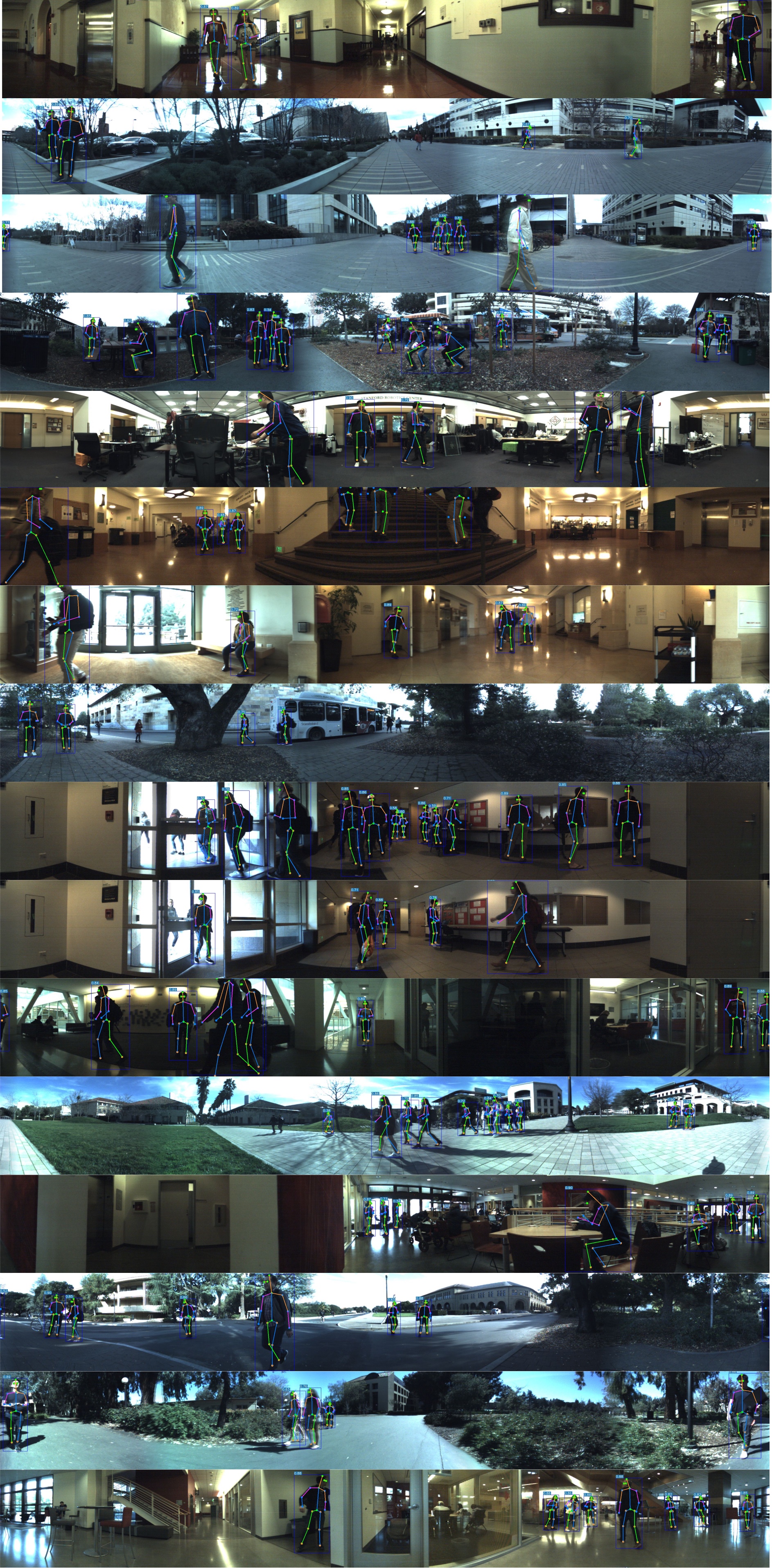}
    \vspace{-.5em}
    %  \vspace{-0.5em}
    \caption{Visualization of Yolo-Pose predictions on 27 testing sequences in JRDB-Pose dataset (cont.)}
    \vspace{-1em}
    \label{fig:testsetviz16}
\end{figure}

We show predictions made by Yolo-Pose, our best performing pose estimation method, on frames from each of the testing sequences in \cref{fig:testsetviz11} and \cref{fig:testsetviz16}. The model used for visualization was initialized from scratch. The predictions are accurate for most poses, demonstrating that JRDB-Pose is sufficiently large enough for the model to learn to predict a wide distribution of poses, even without pre-training.


\begin{table*}[!ht]\centering
\footnotesize
\resizebox{\linewidth}{!}{%
\begin{tabular}{cccccccccccc}\toprule
\multirowcell{2}{\textbf{Pose Estimation} \\ \textbf{Method} {\scriptsize(Training)}} &\multirowcell{2}{\textbf{Tracking} \\ \textbf{Method}} &\multirow{2}{*}{MOTA $\uparrow$} &\multirow{2}{*}{IDF1$\uparrow$} &\multirow{2}{*}{IDSW$\downarrow$} &\multirow{2}{*}{\ospatrack$\downarrow$} &\multicolumn{2}{c}{Components} &\multicolumn{3}{c}{\ospatrack$\downarrow$ by Visibility} \\\cmidrule{7-11}
& & & & & &Card$\downarrow$ &Loc$\downarrow$ &V$\downarrow$ &O$\downarrow$ &I$\downarrow$ \\\cmidrule{1-11}
\multirowcell{3}{Yolo-Pose\cite{maji2022yolo} \\ \scriptsize(COCO only)} & ByteTrack\cite{bytetrack} & 51.00 & 43.63 & 5160 & 0.914 & 0.815 & 0.099 & 0.910 & 0.912 & 0.911 \\
 & UniTrack\cite{unitrack} & 45.74 & 41.35 & 4475 & 0.940 & 0.877 & \textbf{0.064} & 0.938 & 0.939 & 0.938 \\
 & OCTrack\cite{ocsort} & \textbf{55.51} & \textbf{45.33} & \textbf{3906} & \textbf{0.895} & \textbf{0.766} & 0.129 & \textbf{0.892} & \textbf{0.894} & \textbf{0.892} \\
\cmidrule{1-11}
\multirowcell{3}{Yolo-Pose\cite{maji2022yolo} \\ \scriptsize(JRDB-Pose only)} & ByteTrack\cite{bytetrack} & 54.33 & 39.84 & 4730 & 0.920 & 0.796 & 0.124 & 0.917 & 0.920 & 0.917 \\
 & UniTrack\cite{unitrack} & 55.90 & 44.32 & 4206 & 0.928 & 0.849 & \textbf{0.079} & 0.924 & 0.928 & 0.925 \\
 & OCTrack\cite{ocsort} & \textbf{61.28} & \textbf{48.08} & \textbf{3296} & \textbf{0.861} & \textbf{0.692} & 0.169 & \textbf{0.856} & \textbf{0.862} & \textbf{0.855} \\
\cmidrule{1-11}
\multirowcell{3}{Yolo-Pose\cite{maji2022yolo} \\ \scriptsize(COCO$\rightarrow$ \\ \scriptsize  JRDB-Pose)}  & ByteTrack\cite{bytetrack} & 57.69 & 43.68 & 4333 & 0.910 & 0.791 & 0.120 & 0.909 & 0.911 & 0.907 \\
 & UniTrack\cite{unitrack} & 59.37 & 46.82 & 3779 & 0.921 & 0.841 & \textbf{0.080} & 0.919 & 0.921 & 0.918 \\
 & OCTrack\cite{ocsort} & \textbf{63.02} & \textbf{49.04} & \textbf{3394} & \textbf{0.870} & \textbf{0.715} & 0.155 & \textbf{0.867} & \textbf{0.871} & \textbf{0.865} \\
\bottomrule
\end{tabular}
}\vspace{-.5em}
\caption{Multi-person pose tracking baselines evaluated on JRDB-Pose stitched camera images.} \vspace{-1em}
\label{tab:results_tracking_stitched}

\end{table*}

\end{document}